\documentclass[sn-basic,iicol]{sn-jnl}

\jyear{2023}%

\theoremstyle{thmstyleone}%

\theoremstyle{thmstyletwo}%

\theoremstyle{thmstylethree}%

\usepackage{textcomp}
\usepackage{microtype}
\usepackage{booktabs}
\usepackage{times}
\usepackage{algpseudocode}
\usepackage{latexsym}
\usepackage{amssymb}
\usepackage{amsmath}
\usepackage{graphicx}
\usepackage{url}
\usepackage{algorithm}
\usepackage{xcolor}
\usepackage{soul}
\usepackage{array}
\usepackage{bm}
\usepackage{breqn}
\usepackage{lipsum} 
\usepackage{multirow}
\usepackage{enumitem}
\usepackage{natbib}

\usepackage[inline]{trackchanges}
\addeditor{YJ}

\newcommand{\longname}{Synopses of Movie Narratives}
\newcommand{\boldlongname}{\underline{Sy}nopses of \underline{Mo}vie \underline{N}arratives}
\newcommand{\shortname}{\textsc{SyMoN}}
\newcommand{\citeA}[1]{\citeauthor{#1} (\citeyear{#1})}
\newcommand{\vav}{\emph{vis-\`{a}-vis}}

\begin{document}

\title[Article Title]{\centering \longname: a Video-Language Dataset for \\ Story Understanding\\ \tiny \hfill \\
\large \url{https://github.com/insundaycathy/SYMON}}

\author[1]{\fnm{Yidan} \sur{Sun}}\email{SUNY0053@e.ntu.edu.sg}

\author[1]{\fnm{Qin} \sur{Chao}}\email{CHAO0009@e.ntu.edu.sg}

\author[2]{\fnm{Yangfeng} \sur{Ji}}\email{yangfeng@virginia.edu}

\author*[1]{\fnm{Boyang} \sur{Li}}\email{boyang.li@ntu.edu.sg}

\affil[1]{\orgdiv{School of Computer Science and Engineering}, \orgname{Nanyang Technological University, Singapore}}

\affil[2]{\orgdiv{Department of Computer Science}, \orgname{University of Virginia}}


\abstract{Despite recent advances of AI, story understanding remains an open and under-investigated problem. We collect, preprocess, and publicly release a video-language story dataset, \boldlongname{} (\shortname{}), containing 5,193 video summaries of popular movies and TV series with a total length of 869 hours. \shortname{} captures naturalistic storytelling videos made by human creators and intended for a human audience. 
As a prototypical and naturalistic story dataset, \shortname{} features  high coverage of multimodal story events and abundant mental-state descriptions. Its use of storytelling techniques cause cross-domain semantic gaps that provide appropriate challenges to existing models. We establish benchmarks on video-text retrieval and zero-shot alignment on movie summary videos, which showcase the importance of in-domain data and long-term memory in story understanding. With \shortname{}, we hope to lay the groundwork for progress in multimodal story understanding.
}


\maketitle

\section{Introduction}\label{sec1}

Stories are complex artifacts that succinctly encode the human experience. The topic of computational story understanding has attracted substantial research interest over the years \citep{emelin2020moral,chambers-jurafsky-2008-even-chains,ferraro2016unified,wang2021joint,chaturvedi2015modeling,kim2019frowning} 
, but multimodal story understanding \citep{MovieQA,agrawal2016sort,kim2017deepstory,Papalampidi_Keller_Lapata_2021} remains an under-investigated problem. A particular challenge is the lack of multimodal datasets that contain naturalistic and complete stories (\vav{} stories created solely for AI and story segments with incomplete dramatic curves) but are sufficiently short to match the compute available at academic research labs. 

\begin{figure}[t]
	\centering
 \includegraphics[width=\linewidth]{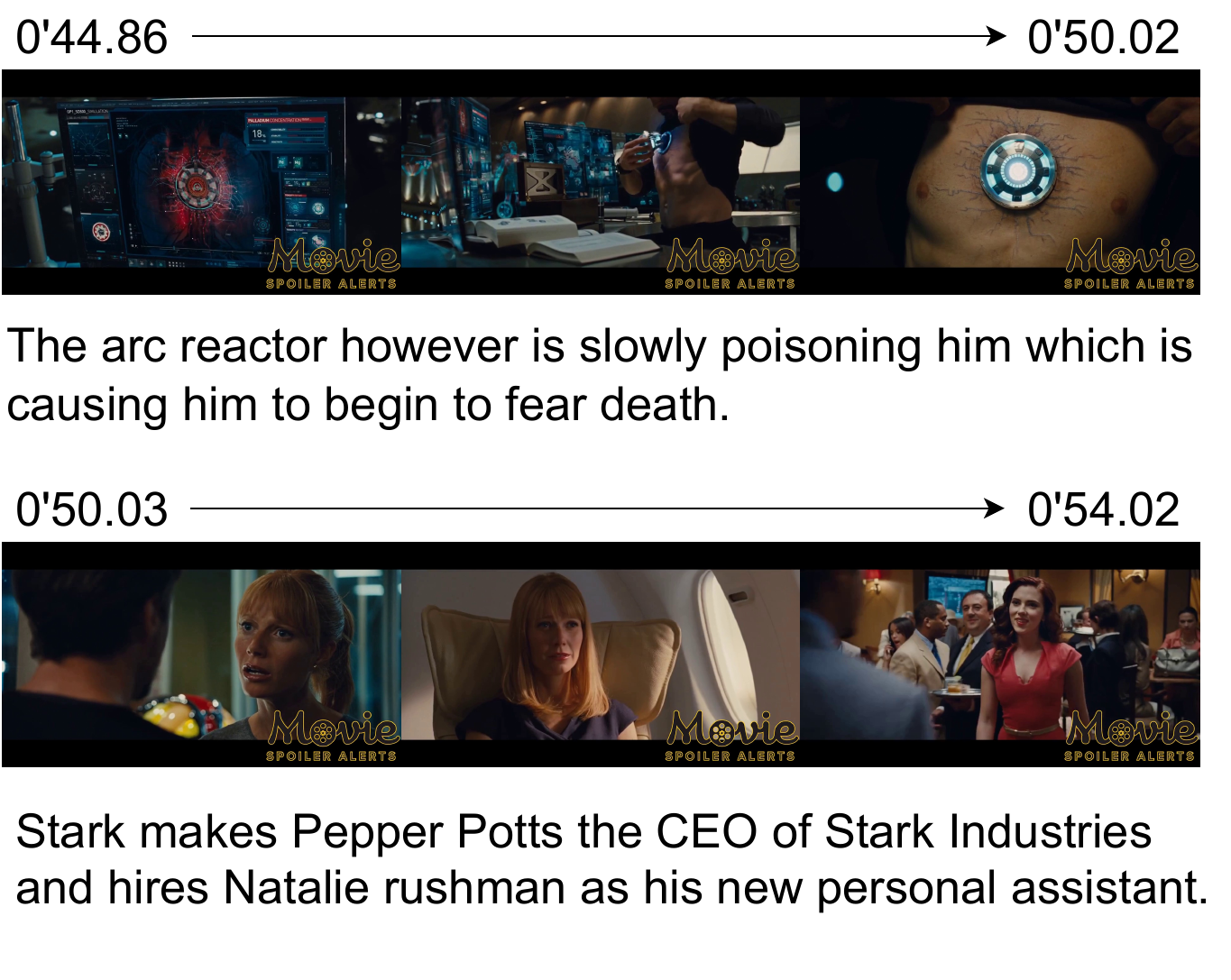}
   \caption{A short video clip from  \shortname{}, shown as a sequence of frames and the narration text. The timestamps indicate starting and ending times. }
   \label{fig:dataset-example}
\end{figure}

In this paper, we present a new video-language story dataset, named \boldlongname{} (\shortname). The dataset is sourced from ``a movie in 5 minutes'' videos on YouTube and contains naturalistic storytelling videos that summarize a movie,  a TV episode, or an entire season of a TV series. The videos are created by human creators for a human audience. They are composed of selected clips depicting key story events and a narrator's retelling of the story, providing condensed yet complete storylines. \shortname{} includes 5,193 user-generated video summaries for a total length of 869 hours. As an illustration of the data, we show one short snippet in Figure \ref{fig:dataset-example}.

\shortname{} is distinguished from other datasets for its extensive coverage of multimodal story events and mental states of story characters as well as an easy-to-process short format. Its contents are representative of storytelling techniques employed in naturalistic multimodal stories. We contend that \shortname{} is a prototypical story dataset that will facilitate the understanding of crucial story elements. 

Although computational story understanding can be traced back to \citep{SAM1975,PAM1983}, today it is a comparably niche research area. 
To explain how \shortname{} facilitates the understanding of crucial story elements, in the following we briefly analyze the story understanding task and the requirements it imposes on datasets.


Converging evidence points to the central role of events and their causal relations in story understanding. First, most definitions of stories in narratology revolve around events and their causal and temporal relations \citep{tomashevsky1925teoriya,Rimmon-Kenan1983,prince2003dictionary,schmid2010narratology,genette2014nouveau}. A well-known definition due to \cite{forster1985aspects} is that ``the king died and then the queen died'' is insufficient but ``the king died and then the queen died of grief'' qualifies as a story (which Forster called a plot) because the latter introduces a causal relation between the two events. Second, cognitive science research indicates causal relations are key to human story understanding \citep{Graesser1991QuestionAI,Trabasso1985}. Third, early works in story generation often assemble events into a causal network \citep{Meehan-TaleSpin-1976,young1994decomposition,Riedl-intentions-2010}. 

\begin{figure}[t]
	\centering
 \includegraphics[width=\linewidth]{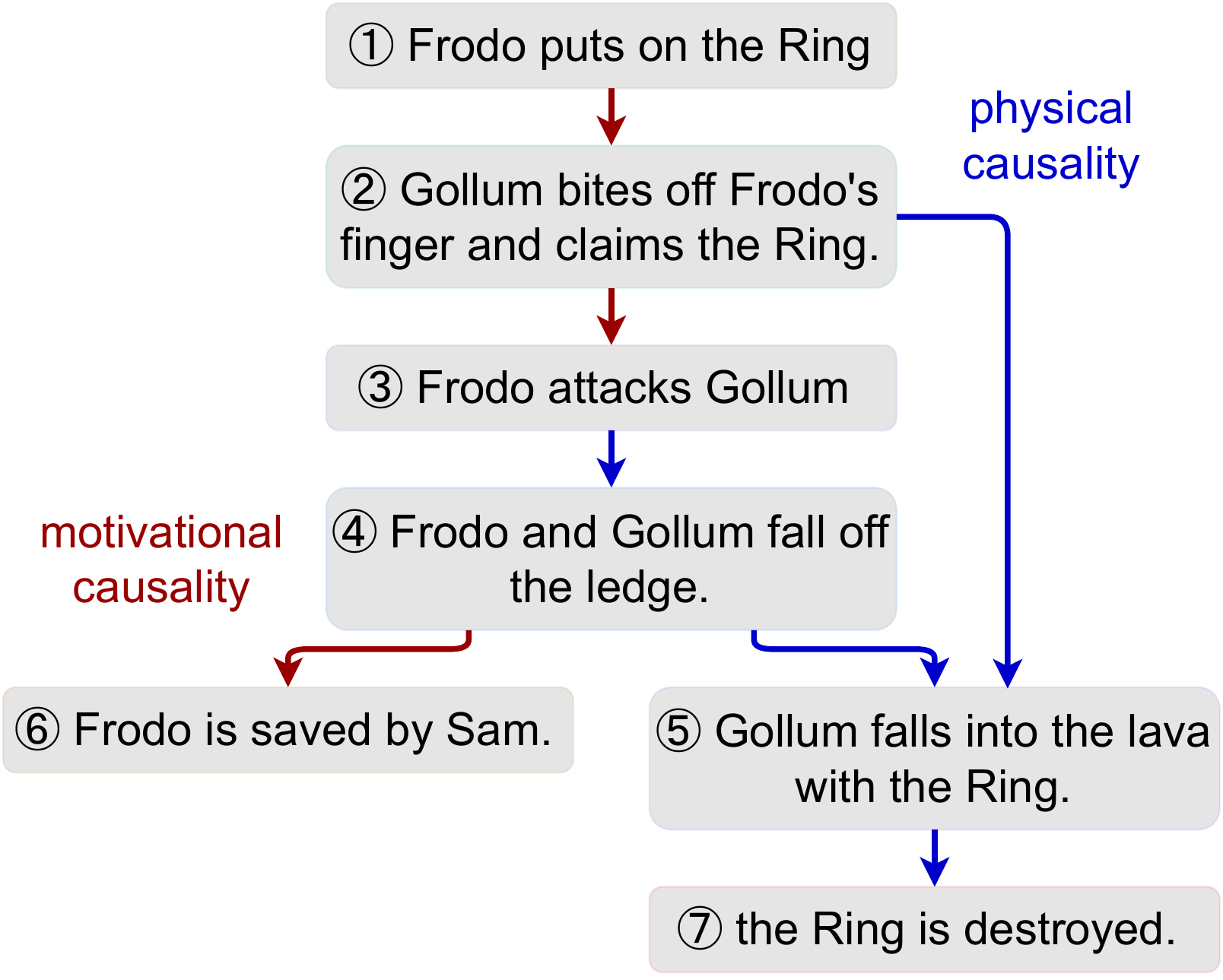}
   \caption{A event graph of causal connections between 7 events from \emph{The Lorad of the Rings}. }
   \label{fig:causal_graph}
\end{figure}

In Figure \ref{fig:causal_graph}, we show a causally related network of events from \emph{The Lord of the Rings: The Return of the King}. Following \cite{trabasso1989logical}, we differentiate causal relations as either physical or motivational. Physical causality is mediated by physical laws, whereas motivational causality involves the voluntary actions of an individual caused by internal motivation, which may be reaction to external stimuli. For example, Gollum's desire for the Ring causes him to bite off Frodo's finger. 

An event can be described at several different levels of granularity; choosing the right level is important when building such event networks. Too much detail could hinder the identification of important events that drive the plot. For instance, in the LSMDC dataset, the event ``Gollum bites off Frodo's finger and claims the Ring'' is subdivided into Gollum lifting his hand, Gollum biting, Frodo screaming, Gollum celebrating, Gollum holding the Ring, etc (Figure \ref{fig:textual_granularity}), potentially causing a computational system to be overwhelmed and lose the forest for the trees. On the other hand, too little detail is also detrimental. 
Coincidentally, cognitive research indicates the existence of a privileged level of event granularity that strikes the right balance between efficiency and information loss and is dominant in human cognitive processing \citep{barker1955midwest,abbott1985representation,rosch1978principles,zacks2001event}.

We aim to empirically verify if \shortname{} portrays stories at the privileged level of event granularity. 
Nevertheless, it is difficult to determine the right granularity of event \emph{a priori}, as event descriptions are also moderated by the story context. For example, if the story is about what happens after the Ring is destroyed, the fight preceding its destruction may be abridged or omitted. 
Instead of manually coding event granularity, which would be highly subjective, we adopt human-written story summaries from Wikipedia\footnote{\url{https://github.com/markriedl/WikiPlots}} as gold references, which the cognitive science theory would posit to be at the privileged level. In Section~\ref{sec:story_coverage}, we show that \shortname{} contains significantly more overlap with the Wikipedia summaries than similar multimodal movie story datasets. This also makes \shortname{} more likely to contain complete dramatic curves \citep{li-etal-2018-annotating} than other datasets.

Another advantage of \shortname{} is that it contains abundant descriptions about mental states of story characters. 
As shown in Figure \ref{fig:causal_graph}, motivational causality plays an important role in relating events. Character emotions and intentions help in understanding motivation causality \citep{Trabasso1989,Riedl-intentions-2010}. For example, in Figure~\ref{fig:dataset-example}, Tony Stark's fear of death motivates him to make arrangements for his company. 
Experimentally, we show that \shortname{} describes mental states of story characters, such as their emotions and intentions, much more frequently than comparable datasets (see Section~\ref{sec:mental-state}).

Yet a third characteristics of \shortname{} is the use of storytelling techniques in \shortname{} causes substantial disparity in the semantic content of the visual and textual modalities, which we refer to as the cross-modality semantic gap (see Section \ref{sec:semantic-gap}). These techniques, such as omission of events easily inferred from context and the use of symbolic and indirect visual representation, are widely employed in multimodal storytelling but are rarely covered in traditional video-text datasets. Through a video clip sequencing experiment and a principled probabilistic analysis, we estimate the cross-modality semantic gap of \shortname{} to be about 31.4\%, whereas comparable datasets have gaps that are either too large to bridge or too small to provide a meaningful challenge. Hence, we argue that  \shortname{} is uniquely suited to computational understanding of multimodal storytelling techniques. 

We propose cross-modal retrieval and sequential cross-modal alignment as evaluation tasks for story understanding on \shortname. Correctly matching the video clips and text snippets of \shortname{} requires understanding of the story semantics in order to bridge the aforementioned cross-modality semantic gap. Moreover, retrieval and alignment tasks can be objectively evaluated and annotated, whereas annotating storytelling techniques would require substantial subjective judgments. Together, the weakly supervised \shortname{} and the fully annotated YMS dataset \citep{nmatch} form a complete test suite and a new challenge for the multimodal research community.

As a baseline for future research, we present a simple neural network equipped with a history retrieval mechanism for tackling the story understanding tasks. Results indicate that modeling long-range dependencies provides benefits for story understanding.

In summary, we make the following contributions in this work:
\begin{itemize}
\itemsep0em 
    \item We collect, preprocess, and publish a large-scale movie summary dataset, \shortname. We demonstrate that large-scale and in-domain data boost performance on movie understanding tasks. 
    \item We experimentally characterize the \shortname{} dataset in three aspects crucial for story understanding: (1) event descriptions at the right level of granularity and story completeness. (2) the amount of mental-state descriptions, and (3) the semantic divergence between text and video. The experiments underscore the differences between \shortname{} and existing datasets. 
    \item To facilitate future research, we establish baselines for bi-directional text-video retrieval on \shortname{} and zero-shot video-text alignment on YMS. We demonstrate the effectiveness of modeling long historical context in story understanding. 
\end{itemize}

\section{Background and Related Work}
\label{sec:background}

\subsection{Background on Story and Event Structures}
\label{sec:narratology-cogsci}

The structuralist school of narratology has offered different variations for the definition of a narrative or a story \citep{prince2003dictionary}, but most theorists regard events and their relations to be key to story structures. \cite{forster1985aspects} argued that a minimal story must contain two events that are causally related. \cite{tomashevsky1925teoriya} argued for the necessity of causality whereas \cite{genette2014nouveau} believed one event alone can constitute a story. \cite{schmid2010narratology} pointed out that it can be difficult to ascertain the existence of causal relations and suggested temporal relations should suffice. It follows that a story understanding dataset should offer opportunities to understand events and their relations. 

Given the importance of events in story understanding, we need to ask how detailed should the event descriptions be. One event can be described as a whole or as several parts. For example, the event ``going to school'' has parts like ``getting dressed'', ``waiting for the school bus'', ``getting on the bus'' and so on, and each part can be further decomposed. Such part-whole relations form an event partonomy, a cognitive structure that plays important roles in narrative comprehension and event understanding \citep{zacks2001event,sridhar2010relational}. 

Cognitive science research shows there is a privileged level in the event partonomy that people naturally adopt when talking and thinking about events \citep{barker1955midwest,abbott1985representation,rosch1978principles,1978Scripts,zacks2001event}. 
The right level in the event partonomy offers enough information at reasonable efficiency. Therefore, we argue that a good story understanding dataset should describe events at this privileged level.



It is also worth mentioning the structuralist perspective of the story arc, or the constituents of a complete story. \cite{Aristotle-Poetics} stated that a story contains a beginning, a middle and an ending. \cite{Freytag1863} proposed a five-element structure consisting of the exposition, rising actions, the climax, falling actions, and the denouement. The most important variable in Fretag's structure is dramatic tension, which rises to its climax through rising actions before falling to the denouement. Several prominent variations of story arcs exist and \cite{li-etal-2018-annotating} attempted to offer a unifying annotation scheme. Inspired by these theories, we argue that a good story understanding dataset should contain complete stories that cover as much of the story arc as possible.

\subsection{Background on Emotion, Motivation, and Intention}
\label{sec:background-mental-state}

The creation of story characters with realistic motivation and behaviors is a hallmark of storytelling excellence \citep{Burgess2022,Murakami2022}. Cognitive studies indicate that emotions play a central role in human motivation and decision making \citep{naqvi2006role}. More specifically, the cognition-appraisal-coping framework \citep{smith1990emotion,gratch2004domain} posits that external stimuli are processed by various cognitive processes, whose results are appraised for potential emotional arousal. Some emotions, especially negative ones, require regulation and coping; these emotions motivate either external behaviors that attempt to change the environment, or psychological adaptation of the person's own stances and relations with the external world.

Due to space limitations, the above discussion of emotion and motivation is vastly simplified. But it is apparent that emotions, motivations, and intentions of story characters are a crucial aspect of stories. Indeed, \cite{Trabasso1989} proposed motivation as one type of causal relations in stories; \cite{Riedl-intentions-2010} introduced motivations and intentions into plan-based story generation. Several text-based datasets describe this aspect of stories \citep{rashkin-etal-2018-modeling,rashkin-etal-2018-event2mind,sap2019atomic}. Therefore, we argue that a good story understanding dataset should facilitate computational understanding of how emotions of the story characters create intentions and motivate behaviors.

\subsection{Datasets for Event and Story Understanding.}
Existing datasets annotate the temporal aspects of events, such as temporal precedence and duration \citep{uzzaman-etal-2013-semeval,chambers-etal-2014-dense,ning2020torque,zhou-etal-2021-temporal,vashishtha-etal-2019-fine,vashishtha-etal-2020-temporal}, and causal relations between events \citep{ogorman-etal-2016-richer,Roemmele-COPA-2011}.  

Several datasets explore individual components of stories, including sentence ordering \citep{gangal2021nareor}, social norms and moral consequences \citep{emelin2020moral}, plausible antecedent events \citep{Brahman2021}, intentions and effects on mental states \citep{rashkin-etal-2018-event2mind}, high-level story structures \citep{ouyang-mckeown-2015-modeling,li-etal-2018-annotating}, 
and story character descriptions \citep{Brahman2021}. \cite{sap2019atomic} consider relations between events, persona, and mental states.
Some datasets aim at summarization of screenplays or conversation transcripts \citep{gorinski-lapata-2015-movie,papalampidi-etal-2020-screenplay,SummScreen-2021}. 

Some QA datasets are conditioned on comprehension of story texts, such as MCTest \citep{richardson-etal-2013-mctest}, NarrativeQA \citep{NarrativeQA-2018}, and FriendsQA \citep{yang-choi-2019-friendsqa}. Multimodal counterparts include MovieQA \citep{MovieQA}, TVQA \citep{lei2019tvqa}, and Pororo \citep{kim2017deepstory}. However, not every question in the QA datasets requires in-depth narrative understanding. 
\begin{table*}
\caption{Comparison of video description datasets with story content.}

\centering
\resizebox{\linewidth}{!}{
\begin{tabular}{@{}m{8.3cm}llll@{}}
\toprule
  & \begin{tabular}[c]{@{}l@{}}Video \\ hours\end{tabular} & \#Videos (\#Clips)  & \#Sent & Vocab.   \\
\midrule
CMD \citep{bain2020condensed}        & 1,270  & 3,605 (33,976)   & 35,681    & 15,272       \\
%
MovieNet \footnote{video release pending at the point of writing this paper}  \citep{huang2020movienet}  & 2,000  & 1,100  &         &            \\
LSMDC \cite{rohrbach2017movie}  & 147   & 200 (128,085)  & 128,118 & 22,500       \\
%
Pororo  \citep{kim2017deepstory}   & 20.5   & 171 (16,066)   & 43,394  &              \\
MovieGraph \citep{moviegraphs} & 94.0   & 51 (7,637)      & 20,849  &   \\   
   
\shortname{} (Ours)    & 869   & 5,193     & 683,611 & 40,116   \\       
\bottomrule
\end{tabular}}

\label{tab:statistics}
\end{table*}

\subsection{Video-Text Movie Story Datasets.}
A number of datasets supply story content extracted from movies. The Large-Scale Movie Description Challenge (LSMDC) \citep{rohrbach2017movie} provides meticulously detailed language descriptions of the video content initially intended for the visual impaired. Although these descriptions are highly accurate, they are not representative of real-world storytelling. 

YouTube Movie Summary (YMS) \citep{nmatch} contains 94 YouTube movie summary videos with human-narrated storylines. The Condensed Movies Dataset (CMD) \citep{bain2020condensed} gathers 7 to 11 key clips from each movie with one-sentence descriptions for each clip. Pororo \citep{kim2017deepstory} captures 20-minute cartoon episodes, in-show conversations, and human-written descriptions. MovieNet \citep{huang2020movienet} annotates 2000 hours of movies with extensive annotations and aligned movies scripts. However, due to copyright considerations, at the time of writing, the video release is still pending.  

Other types of video annotations have been explored, including 
semantic roles and event relations \citep{vidsitu}, character relationships and types of speech \citep{wu2021towards}, and movie graphs \citep{moviegraphs}.

In this work, we collect a large-scale, readily available, multi-reference dataset of human-curated movie summaries, \shortname{}. The dataset can be leveraged for various story understanding and generation tasks such as sequential text localization, story generation from video, and movie summarization. To our knowledge, \shortname{} is the largest dataset for short naturalistic storytelling videos.

\section{Collection and Preprocessing}
In this section, we describe the procedures for collecting and preprocessig of the \shortname{} dataset, and also report some data statistics. 

\subsection{Dataset Collection and Statistics}
\label{sec:datsetstat}
We apply the following procedure for data collection. First, we manually identify relevant YouTube channels by searching with keywords such as ``movie summary'', ``movie recap'', and ``movie shortened''. We download all videos from the identified channels and accompanying subtitles, which may be written by humans or automatically generated by YouTube. In subsequent processing, we do not distinguish between manual and automatic transcriptions, but we include a label in the dataset for automatic transcript.
Videos without subtitles are excluded. Finally, we perform rule-based extraction of movie names from metadata and subtitles and discard videos that are not movie summaries. 

This yields a total of 5,193 videos with an average length of 9.5 minutes and a total length of 869 hours. On average, the narration in one video contains 1,717 words or 131 sentences. The overall vocabulary size is 40,116. \shortname{} contains summaries for 2,440 movies and TV series, of which 432 have more than 1 summary. The most popular movie, \emph{The conjuring}, has 12 summaries. On average, one movie or TV series in the 432 has 2.79 summaries. 
Compared to existing movie story datasets (see Table \ref{tab:statistics}), \shortname{} has the largest collection of short storytelling videos as while as the largest vocabulary. 


\subsection{Preprocessing}
\label{sec:preprocessing}

\noindent \textbf{Subtitle Masking.} 
Some videos have subtitles embedded in the video. In tasks like text-to-video retrieval, the embedded subtitles may become a shortcut feature, causing neural networks to rely on only optical character recognition. 

To eliminate shortcuts, we locate embedded subtitles and mask them out. 
For efficiency, we randomly sample 100 frames from each video and apply an accurate text detection technique \citep{baek2019character}. Observing that the subtitles are almost always at the same location in a single video, we  
take the minimum bounding box that can cover all embedded subtitles in all 100 frames as the masked region; we set all pixels in the region to black. 

\vspace{0.1in}
\noindent \textbf{Punctuation Restoration.} 
We acquire subtitle texts from YouTube directly. Sometimes the texts are the result of automatic speech recognition, which cannot recognize punctuation. To fix this, for every unpunctuated narration text, we generate punctuation with the approach proposed by \cite{alam2020punctuation}. 

\vspace{0.1in}
\noindent \textbf{Scene Segmentation.}
\label{sec:segmentation}
Later experiments require temporal segmentation of videos based on camera cuts. For this purpose, we run the dataset through the network of \citep{souvcek2020transnet}, which detects hard camera cuts. A scene, defined as a continuous shot between two cuts, lasts 2.2 seconds on average. 
This is similar to CMD, another movie dataset, whose scenes last 2.4 seconds on average. However, average scenes in ActivityNet \citep{caba2015activitynet} and Kinetics-400 \citep{kay2017kinetics} last for 11.1 seconds and 30 seconds respectively. This shows camera cuts in movies are much more frequent than the user-generated videos in ActivityNet and Kinetics. 

\section{Dataset Analysis: Event Granularity and Story Coverage}
\label{sec:story_coverage}
As discussed in Section \ref{sec:narratology-cogsci}, to assist an AI system in acquiring the ability to understand stories like humans, it is desirable for the dataset to represent events at a level of granularity akin to human cognitive representation. This level of granularity should be neither too concise nor too detailed. From the perspective of event causality, the right level of event granularity should omit details that do not advance the storyline. In addition, it is desirable for the dataset to contain complete stories spanning the entire story arc. 

In this section, we argue that \shortname{} provides event descriptions at the right level of granularity and relatively complete stories. We first present a qualitative comparison between three video-text story datasets. After that, we quantify the similarity between human written story summaries from Wikipedia and textual descriptions of video-text story datasets. 

\subsection{Qualitative Example}
\begin{figure*}[p]
	\centering
   	\makebox{\includegraphics[width =\linewidth]{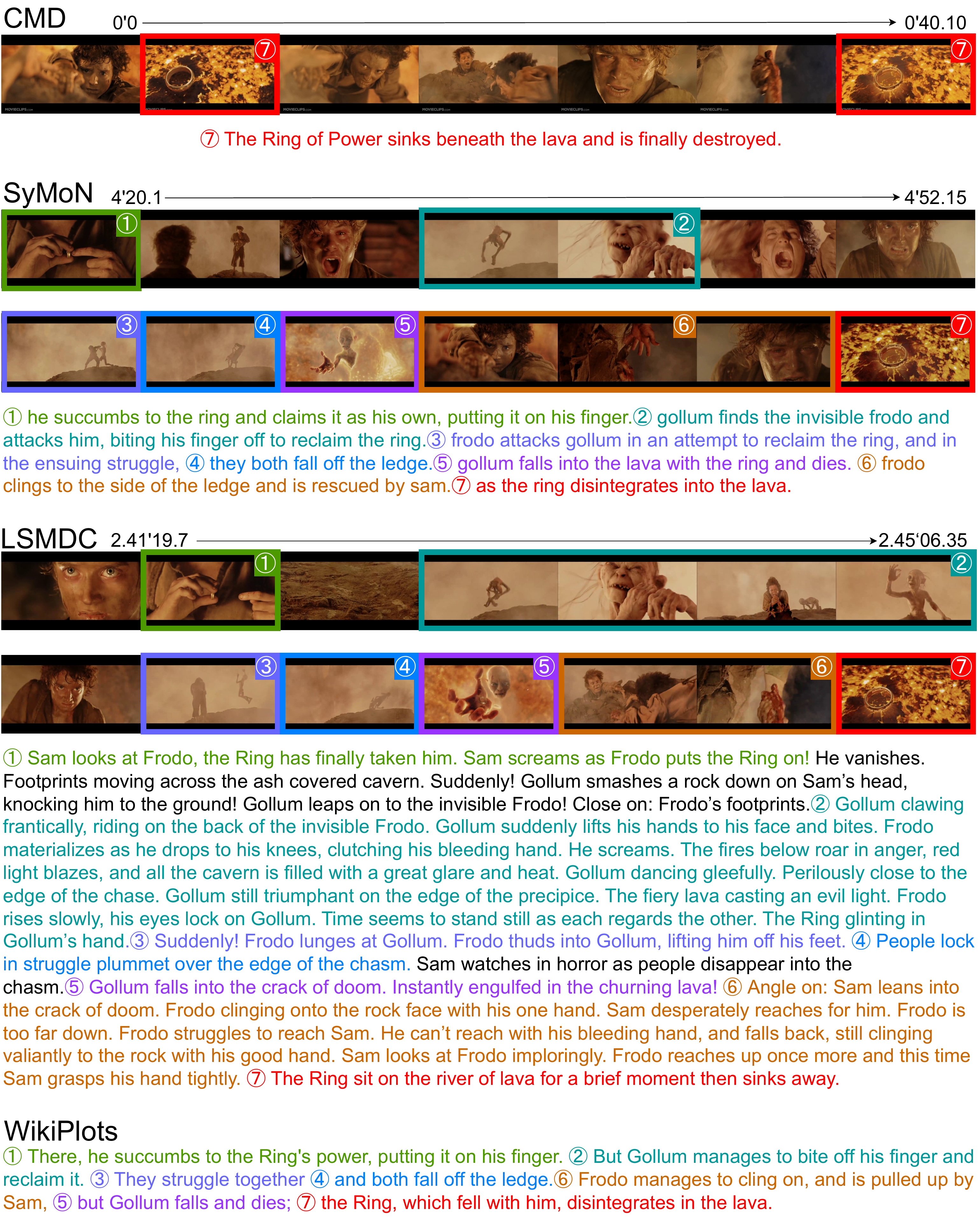}}
    \caption{Compare textual granularity of CMD, LSMDC \shortname{}, and WikiPlots. The frames are respectively sampled from the \shortname{}, CMD, and LSMDC video summaries of the movie \emph{The Lord of the Ring: The Return of the King}. Video-text correspondence are manually marked by colors and numbers. We present the WikiPlots summary of the same movie segment as the last section. 
   }
   \label{fig:textual_granularity}
\end{figure*}

Figure~\ref{fig:textual_granularity} compares \shortname{} with two other video-text story datasets, CMD and LSMDC, as well as WikiPlots story summaries, using the same scene taken from \emph{The Lord of the Rings: The Return of the King}. We identify 7 major plot events that form a sub-story arc with rising and falling tension:
\begin{itemize}
    \item Frodo puts on the Ring. (rising action)
    \item Gollum bites off Frodo's finger and claims the Ring. (rising action)
    \item Frodo attacks Gollum. (rising action)
    \item Frodo and Gollum fall off the ledge. (climax)
    \item Gollum falls into the lava with the Ring. (climax)
    \item Frodo is saved by Sam. (falling action)
    \item The Ring is destroyed. (falling action)
\end{itemize}

Despite having the shortest video run time (31 seconds), \shortname{} covers all 7 events. In comparison, the CMD text describes the final outcome, the destruction of the ring, but omits the rest completely. The gap between the brief CMD text and the 40-second video is hence substantial. The third dataset, LSMDC, is designed to help the audience with visual impairment. As a result, its text offers a blow-by-blow recount of the video. Although LSMDC texts match well with the video and may be useful for understanding individual shots, excessive details may complicate story understanding at the event level and obscure the big picture. For example, the description ``Gollum suddenly lifts his hands to his face and bites'' and subsequent sentences render it difficult to infer that Gollum reclaims the ring this way, whereas in WikiPlots and \shortname{} this is explicitly explained. 

From a cognitive science perspective (see Section \ref{sec:background}), LSMDC descriptions sit at a lower level on the event partonomy than \shortname, and CMD descriptions sit at a higher level. \shortname{} has the best alignment with the human-written summary from WikiPlots, which we argue is at the privileged level of event partonomy that balances information and efficiency. In the next section, we verify this empirically.

\subsection{Quantitative Assessment of Story Coverage}
We computationally estimate the extent to which stories in CMD, LSMDC, and \shortname{} cover the sentences in WikiPlots. We use a three-step procedure for computing story coverage. First, we match movie summary in our dataset to their WikiPlots summaries by name\footnote{For story coverage evaluation on \shortname{}, only movies whose plot summaries can be found in WikiPlots is included.}.

Second, we estimate if a sentence from the video narration is equivalent to a sentence in WikiPlots using a distance metric. 
We experiment with two distance metrics between input sentences $a$ and $b$. The first metric is the average probability of $a$ entailing $b$ and $b$ entailing $a$ from the natural language inference classifier of \cite{nie-etal-2020-adversarial}. Alternatively, from Sentence-Bert (SBERT) \citep{reimers2019sentence}, we compute the cosine distance between the sentence embeddings. 

Finally, we find the best correspondence between two texts using Dynamic Time Warping (DTW) \citep{dtw1994}. 
Briefly, DTW utilizes dynamic programming to find the minimum-cost correspondence between two sequences, the WikiPlots sentence sequence $A$, and the narration sentence sequence $B$. 
We refer readers to Appendix \ref{appendix:coverage} for details of the DTW algorithm. DTW has two model parameters, $\delta_A$ and $\delta_B$, which denote the costs for skipping a sentence in sequences $A$ and $B$.
We determine them from manual annotations, as described below. 

We manually labeled 1,000 sentences from CMD for correspondence to Wikiplots stories, 1,000 sentences from \shortname{}, and 2,000 sentences from LSMDC, which amount to 83 texts from CMD, 8 texts from \shortname{}, and 4 texts from LSMDC, due to different text lengths. A second annotator labeled a small portion of data from each dataset to compute inter-rater reliability. Cohen's Kappa on \shortname{}, CMD, and LSMDC are 0.86, 0.59, and 0.33 respectively. We believe the low agreement on LSMDC is caused by the much longer lengths of LSMDC texts than Wikiplots stories, which led to difficulties in annotating precise correspondence. 

On this annotated dataset, we use grid search to find the optimal $\delta_A$ and $\delta_B$ that maximize the accuracy of DTW, which is defined as 
\begin{equation}
\label{eq1}
    	\text{Accuracy} = \frac{1}{2} \left(\frac{{M}^{\text{wiki}}}{{N}^{\text{wiki}}}+\frac{M^{\text{text}}}{N^{\text{text}}}\right).
\end{equation}  
Here ${M}^{\text{wiki}}$ and ${N}^{\text{wiki}}$ are the number of correctly matched and the total number of WikiPlots sentences, respectively. $M^{\text{text}}$ and $N^{\text{text}}$ are the number of correctly matched and the total number of sentences in the video narration. We do not directly optimize alignment percentage because doing so will cause DTW to inflate alignment percentage by matching sentences incorrectly. 

\begin{table}
\caption{Estimated alignment percentage with two distance metrics and Dynamic Time Warping.}
\centering
\resizebox{\columnwidth}{!}{
\begin{tabular}{@{}llll@{}}
\toprule
Distance Metric & CMD     &  LSMDC &\shortname{} \\
\midrule
Symmetric Entailment & 10.6\%  & 18.1\% & 44.3\%\\
SBERT Cosine & 28.7\%  & 38.3\% & 81.2\%\\
\bottomrule
\end{tabular}}
\label{tab:story-coverage}
\end{table}

With the optimal $\delta_A$ and $\delta_B$, we perform DTW again and calculate story coverage as the proportion of WikiPlots sentences matched with narration sentences, 
\begin{equation}
	\text{Coverage} =\frac{1}{K}\sum\limits_{i}^{K}{\frac{M_{i}^{\text{wiki}}}{{{N}^{\text{wiki}}_i}}},
\end{equation}
where the summation is over all WikiPlots summary texts and $K$ is the number of WikiPlots movies appearing in the video dataset. In the $i^{\text{th}}$ WikiPlots text, $M_{i}^{\text{wiki}}$ denotes the number of matched sentences and ${N}^{\text{wiki}}_i$ denotes the total number of sentences.

Table \ref{tab:story-coverage} shows the story coverage results. Though the two distance metrics result in different coverage numbers, the ranking between methods remains unchanged. 
Of the three datasets, \shortname{} provides the highest story coverage. LSMDC takes second place, partially benefiting from the extreme lengths of its texts. The results demonstrate that \shortname{} stories cover more major plot points than the other two datasets. As most WikiPlots summaries are complete stories, and are probably at the privileged level of event partonomy, the findings suggests that \shortname{} contains relatively complete story arcs and and has suitable granularity for story understanding.

\section{Dataset Analysis: Mental-state Descriptions}
\label{sec:mental-state}
As explained in Section \ref{sec:background-mental-state}, characters' mental state such as emotions, intentions, and motivations play crucial roles in the causal network of events and in story understanding. 
In this experiment, we measure the frequency of words related to emotions, intentions, and motivations in the text associated with the videos. For emotional words, we adopt the WordNet-feelings dataset \citep{siddharthan2018wordnet}, which includes 11,387 emotion-related words identified by human experts. For motivation and intention words, we find 200 nearest neighbors of the words ``motivation'' and ``intention'' using 300-dimensional fastText embedding\footnote{Acquired from \url{https://github.com/facebookresearch/fastText/blob/master/docs/crawl-vectors.md}.} \citep{bojanowski2017enriching} trained on Wikipedia and Common Crawl. We stop at 200 words as we find additional neighbors to be mostly irrelevant.

Table \ref{tab:mental-state-words} reports word frequencies for every thousand words in four video-language datasets. We observe that \shortname{} employs mental-state words with the highest frequency. Interestingly, \shortname{} contains  2.5 times as many intention-related words as the next dataset, CMD, whose textual descriptions also focus on story content.
LSMDC, containing literal descriptions of movie clips, is ranked at the third position. As a sanity check of the metric we adopt, we compute the same statistics for ActivityNet Captions \citep{activitynet-captions}, containing matter-of-fact descriptions of actions in generic user-uploaded videos. As we expect, ActivityNet Captions uses the least of mental-state words. Overall, the ranking is consistent with the nature of the datasets, as story text describes mental states more often than literal descriptions of generic videos. \shortname{} appears to be the most prototypical story dataset. 

One possible caveat of the above analysis is that it does not exclude narrator commentary, which may have inflated the word frequencies. We further recruited human annotators from Amazon Mechanical Turk (AMT) to label 36,955 sentences (8.5\% of all text from \shortname{}) as either commentary or non-commentary. The workers annotated 8.9\% of sentences as commentary. Over the remaining 91.1\% or 33,666 sentences, we recompute the frequency statistics and find marginal changes. We conclude that the high frequencies are indeed due to the story content rather than narrator commentary. 
Details of the AMT experiment are in Appendix \ref{appendix:mental-state}. 

\begin{table}
\caption{Frequency of words related to emotion, motivation, and intention per one thousand words.}
\centering
\footnotesize
\resizebox{\columnwidth}{!}{
\begin{tabular}{@{}p{3.2cm}p{0.6cm}p{0.9cm}p{0.9cm}@{}}
\toprule
             & Emotion & Motivation & Intention \\
\midrule
AcitivityNet Captions & 27.5  & 0.51       & 2.7     \\
LSMDC        & 33.5   & 0.62       & 2.8      \\
CMD          & 38.9    & 1.41       & 9.4      \\
\shortname{} (all text)  & 57.6   & 1.58      & 23.9    \\
\shortname{} (excl. commentary) & \textbf{61} & \textbf{1.78} & \textbf{24} \\
\bottomrule
\end{tabular}}

\label{tab:mental-state-words}
\end{table}

\section {Dataset Analysis: Cross-modality Semantic Gap}
\label{sec:semantic-gap}
Multimodal storytelling often utilize the modalities in a complementary manner. For example, \cite{IyyerComics2016} study comics and find that imagery and text are interwoven in the storytelling and understanding any single modality only is insufficient for story understanding. 

Similarly, in \shortname{}, the story narration and the video often are not simple duplicates, but provide complementary information. As a result, the text in \shortname{} often does not exactly match the video content, i.e. cross-modality semantic gap. In this section, we qualitatively analyze the cause and quantitatively estimate the semantic gap.

\begin{figure}[t!]
	\centering
   	\includegraphics[width=\linewidth]{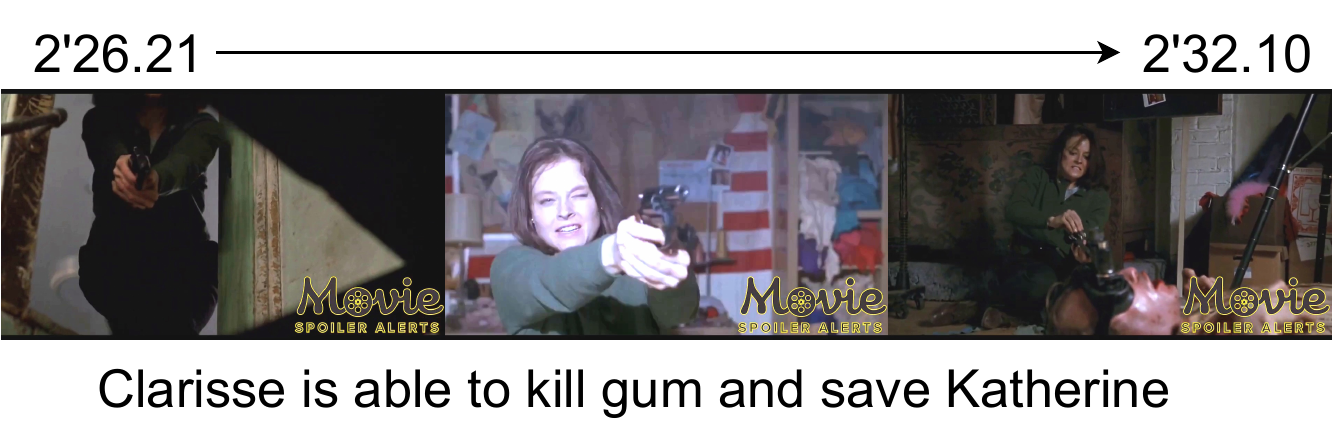}
   \caption{This example shows three frames from \emph{Silence of the Lambs}. The text (kill) describes the effect of the video (shooting).
   }
   \label{fig:recording_bias_effect}
\end{figure}

\subsection{Categories of the Semantic Gap}
\label{sec:semantic-gap-categories}
Through manual examination of the data, we provide a categorization of the mismatch between the narration text and the video. Though the list of categories is likely incomplete, we believe it can provide insight into the types of multimodal storytelling techniques utilized by videos in \shortname. 

\vspace{0.1in}
\noindent \textbf{Moving along the causal chain.}
Instead of describing precisely the event shown in the video, the narrator describes its immediate cause or effect. For example, in Figure~\ref{fig:recording_bias_effect}, the video shows Gum getting shot and then lying on the ground coughing up blood. The text describes the event as ``Clarisse kills Gum.'' The expectation is that the viewer can infer that ``being killed'' is the direct effect of getting shot and therefore can match the text with the video. By utilizing this strategy, the narrator avoids stating the obvious and provides information that complements the video.

\begin{figure}[t!]
	\centering
   	\includegraphics[width=\linewidth]{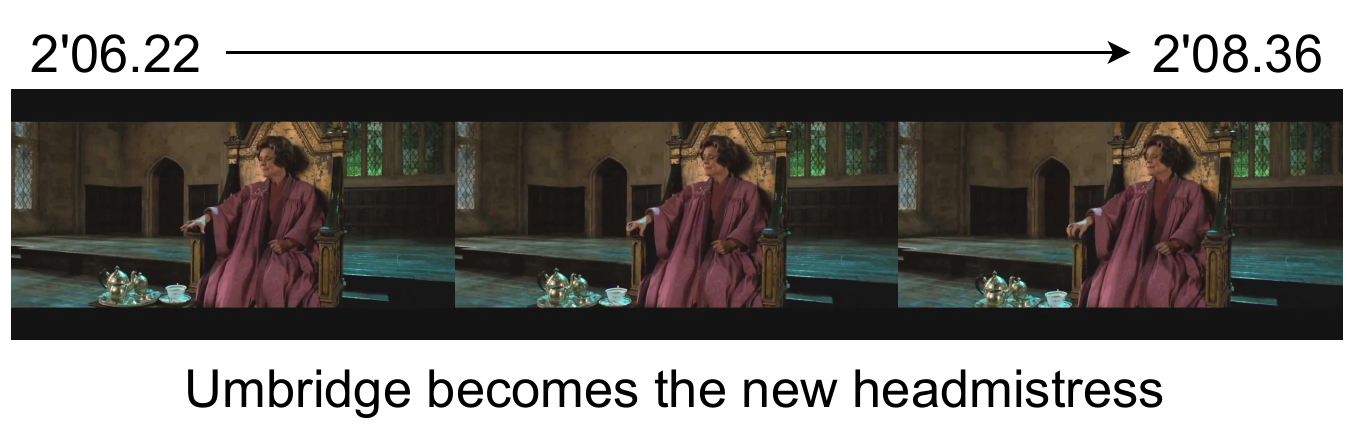}
   \caption{An example from \emph{Harry Potter and the Order of the Phoenix}. A symbolic object, the chair, is used to represent the event Dolores Umbridge becoming headmistress.
   }
   \label{fig:recording_bias_object}
\end{figure}

 \vspace{0.1in}
\noindent \textbf{Symbolic representation of events.}
The storytelling in \shortname{} sometimes employs a symbolic object to represent the occurrence of an event.
For example, in Figure~\ref{fig:recording_bias_object}, the video shows Dolores Umbridge sitting on the headmaster's chair while the narrator states ``Umbridge becomes the new headmistress''. Here, the chair represents the event of Umbridge becoming headmistress. The audience is expected to have commonsense knowledge and be aware of the association between the high-back armchair and someone's status as the headmaster or headmistress.

\begin{figure}[t!]
	\centering
   	\includegraphics[width=\linewidth]{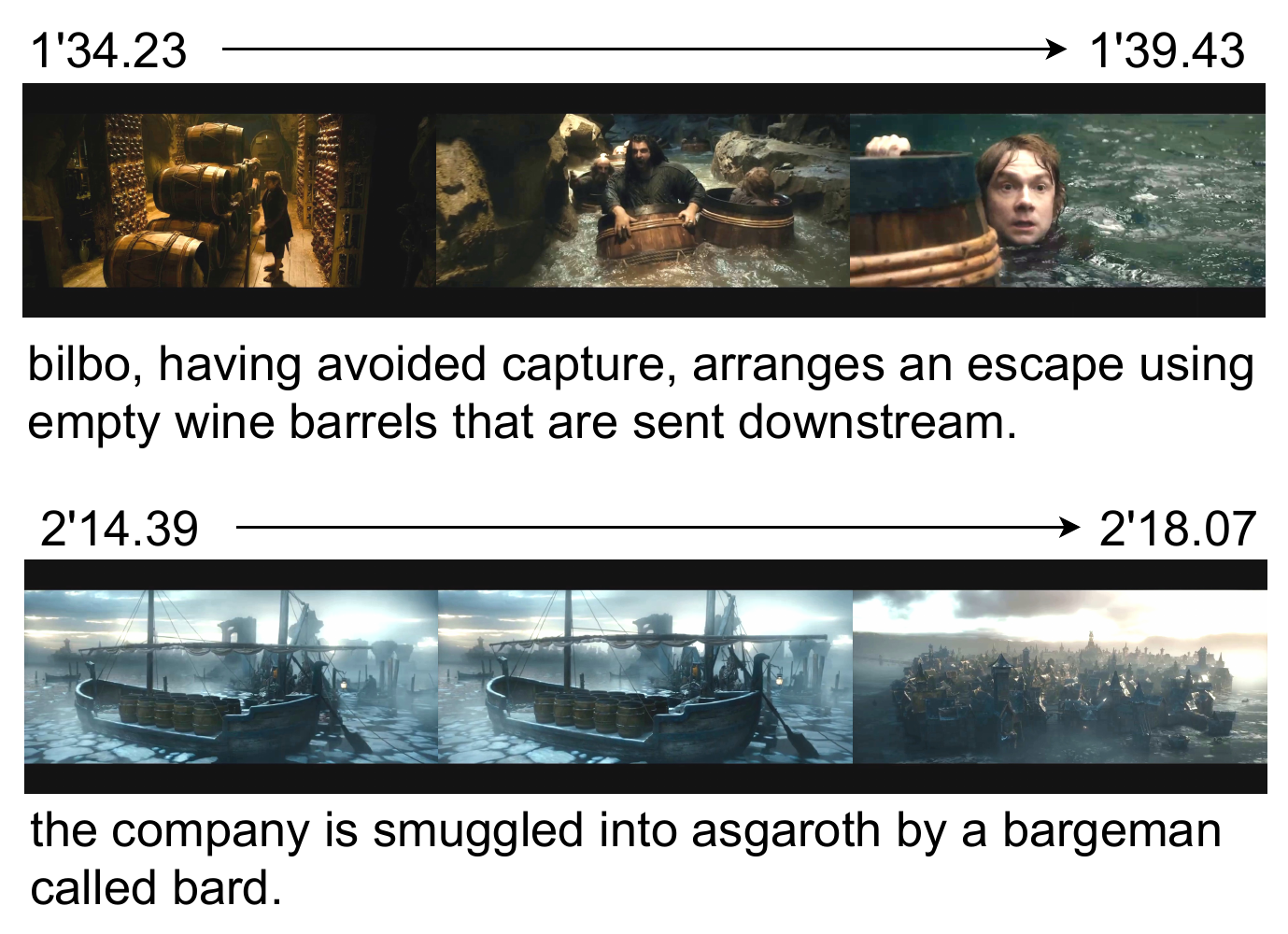}
   \caption{Example from the film \emph{The Hobbit},  The video segment on the first line gives the context (the dwarfs escaping in barrels), which is required to understand the connection between the video and the text on the second line.}
   \label{fig:recording_bias_object_hobbit}
\end{figure}

\vspace{0.1in}
\noindent \textbf{Reference to long-range context.}
Narrator may make references to previous events when explaining the current event, so that long-range context is needed to match current video segment with text. For example, in Figure~\ref{fig:recording_bias_object_hobbit} the video shows a ship full of wine barrels moving toward a town and the text says the company is smuggled into Asgaroth. To understand the connection between video and text, the audience needs to recall earlier in the video, the dwarfs hid themselves in the wine barrels. 

\begin{figure}[t!]
	\centering
   	\includegraphics[width=\linewidth]{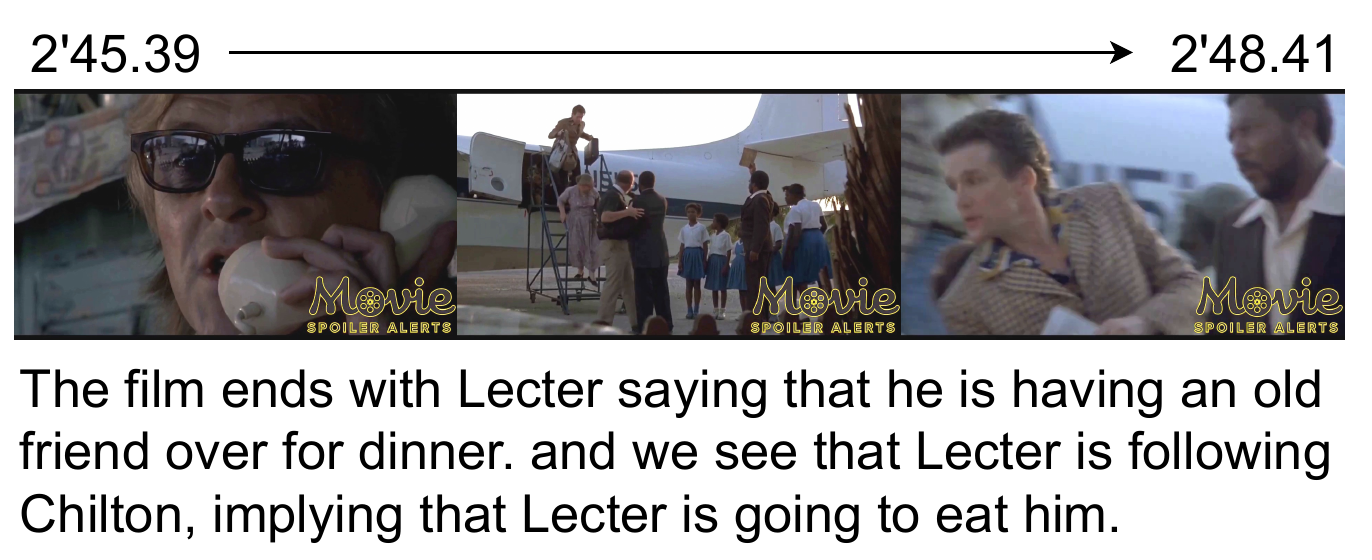}
   \caption{Example from film \emph{Silence of the Lambs}, the text describes an event implied in the movie (Lector eating Chilton). Because there is no matching video of this event the video shows Chilon disembarking a plane.}
   \label{fig:recording_bias_implied}
\end{figure}

\vspace{0.1in}
\noindent \textbf{Explicitly stating the implied.}
The text explicitly explains events that were only implied in the original movie or TV show. In a full-length movie, some events may be omitted and left for the audience to deduce. However, it is much harder to deduce missing events in a short and fast-tempo summary video. Therefore, these events usually need to be explicitly explained in \shortname{}. Since the events are not present in the movie, there is no matching imagery to the narration and the semantic gap is inevitable. In the example shown in Figure~\ref{fig:recording_bias_implied}, the video shows Lector watching Chilton disembarking from a plane. In the movie, the audience knows that Lector is a cannibal and therefore can understand that Lector means to eat Chilton. In the recap video, the event ``implying Lector is going to eat him'' is explicitly stated in the text to help audiences better understand the story.  

\begin{figure}[t!]
	\centering
   	\includegraphics[width=\linewidth]{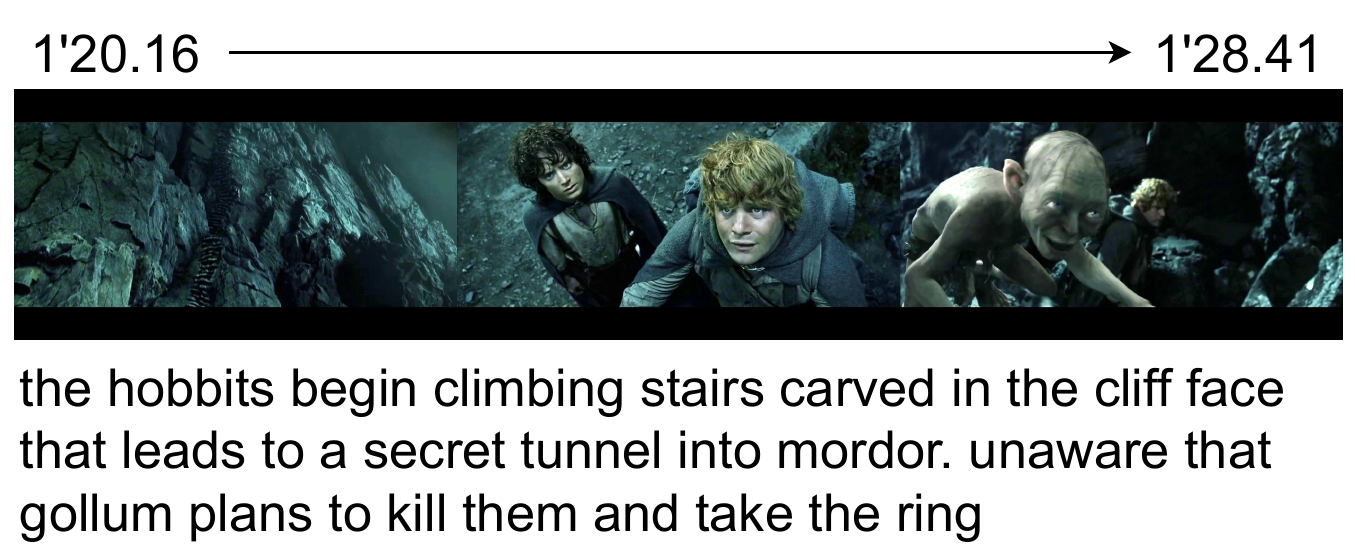}
   \caption{An example from the movie \emph{Lord of the Rings: Return of the King}. The video does not show Gollum's intention but the text explains it to further the story.
   }
   \label{fig:recording_bias_state_of_mind}
\end{figure}

 \vspace{0.1in}
\noindent \textbf{Mental state descriptions.} 
    The narration explicitly describes the mental state of the characters. 
    In a short 5-minute summary video, the characters' emotions, intentions, and motivations need to be directly explained to help audiences understand the story. In the example shown in Figure~\ref{fig:recording_bias_state_of_mind}, the video shows the hobbits climbing stairs with Gollum. Gollum's intention to kill the hobbits and claim the ring is not apparent, but is however crucial for understanding the later events in the story. The narrator makes it explicit. 

\begin{figure}[t!]
	\centering
   	\includegraphics[width=\linewidth]{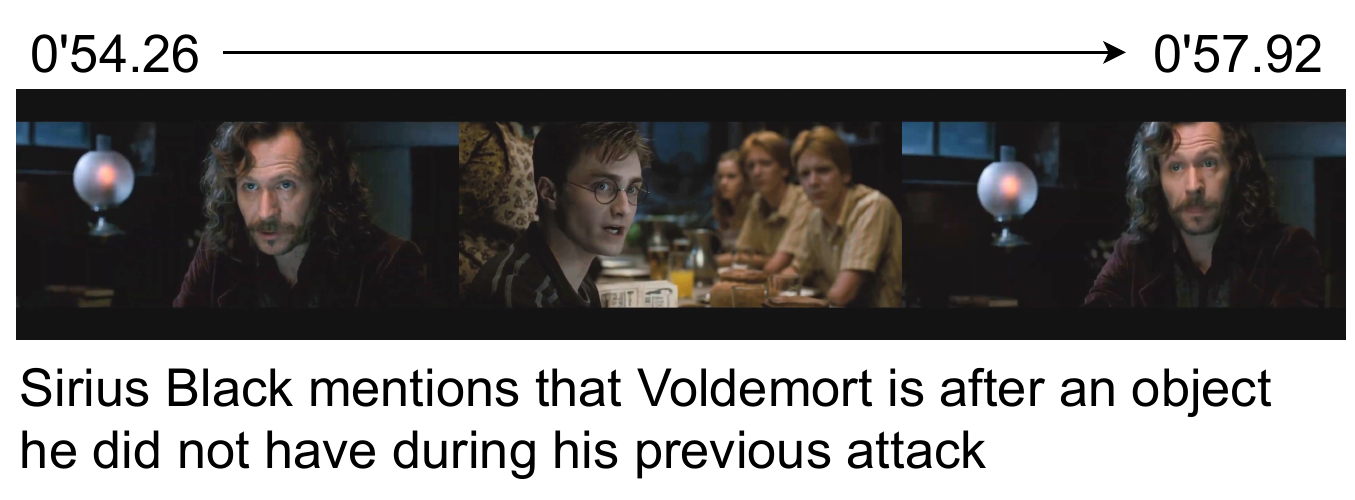}
   \caption{This example shows two frames from the movie \emph{Harry Potter and the Order of the Phoenix}, the video shows two people having a conversation. The text explains the content of the conversation.}
   
   \label{fig:dialogue}
\end{figure}

\vspace{0.1in}
\noindent \textbf{Summarizing dialogues in the original.} 
The narration may summarize the content of a dialogue in the original movie or TV show, while the video shows two people talking. See an example in Figure~\ref{fig:dialogue}. In most videos in \shortname{}, dialogues in the original movie are replaced with a narrated summary. This is because the original dialogue is usually long and contains a lot of information; it is not easy to comprehend the relation between the dialogue and other events in a short amount of time. The summarization omits details irrelevant to the plot and helps the viewer grasp the key content quickly. At the same time, the video often only shows people conversing. The narrator sometimes attempts to show images related to the dialogue content, but for abstract ideas like Voldemort's intention, finding relevant imagery can be challenging, leaving substantial semantic gaps between the two modalities. 

\vspace{0.1in}
\noindent \textbf{Discussion.} 
The examples and analysis above illustrate how videos in \shortname{} deliberately utilize the two modalities in a complementary manner to achieve storytelling goals. This is in stark contrast to most existing visual-language datasets, such as Charades \citep{sigurdsson2016hollywood}, ActivityNet Captions \citep{activitynet-captions}, and LSMDC \citep{rohrbach2017movie}, which are carefully curated or selected so that the two modalities match as closely as possible. 

Such semantic gaps are not merely the result of \shortname{} being short summaries of stories. Rather, most of the causes for the semantic gap we analyzed (omission of events on the causal chain, use of symbolism, reference to past events, etc) are classic film-making techniques ubiquitous to storytelling videos in the wild.  For example, in the famous montage of the \emph{Godfather}, the scene cuts between a series of brutal murders to Michael Corleone standing at an altar as the godfather of Connie's baby. Though never explicitly shown, the audience is expected to infer the cause of the murder is Michael's order. In addition, Micheal's status as the baby's godfather symbolizes the fact that he has become the godfather of a criminal organization. In this sense, \shortname{} provides a representative benchmark for story video comprehension in the wild. Next, we report experiments designed to estimate the extent of video-text correspondence. 



\subsection{Quantifying the Semantic Gap}
\label{sec:coherence_exp}
In this section, we first introduce the task of video segment ordering, whose results enable principled statistical analysis of the semantic gap between the two modalities,  After that, we analyze the performance of neural networks. 

Similar to event / sentence ordering \citep{liu2018graphbased,devlin2018bert}, we predict the correct ordering of two consecutive video segments separated by a hard camera cut. 
This is a binary classification task with two outcomes: video segment 1 preceding segment 2 or vice versa. To create a balanced label distribution, we randomly flip the ordering of the two video segments. We extract the text description that spans the same duration as the two video segments and expand the text to sentence boundaries. From 5,193 (4,733 for training) videos, we generate 136,220 (122,751 for training) data points, each consisting of two video segments and one text snippet.

\begin{table}
\caption{Performance of human annotators on the video segment sequencing task.  }
\centering
{
\resizebox{\columnwidth}{!}{
\begin{tabular}{@{}p{1cm}p{0.5cm}p{0.5cm}p{0.1cm}p{0.1cm}p{0.5cm}p{1.1cm}@{}}
\toprule
 &&&  \multicolumn{2}{c}{Video-text} &  \multirow{2}{*}{Total}&  \multirow{2}{1em}{Estimated Gap}\\
&&& \multicolumn{1}{c}{Correct} & \multicolumn{1}{c}{Wrong} &\\
\midrule
\multirow{3}{*}{\shortname{}}&\multicolumn{1}{l}{\multirow{2}{2em}{Video-only}} & Correct & \multicolumn{1}{c}{75.6\%}          & \multicolumn{1}{c}{7.2\%}  & 82.8\% & \multirow{3}{*}{31.4\%}\\
&\multicolumn{1}{c}{}                           & Wrong & \multicolumn{1}{c}{14.5\%}         & \multicolumn{1}{c}{2.7\%}  & 17.2\% \\ 
\cline{2-6}
&\multicolumn{2}{c}{Total} & \multicolumn{1}{c}{90.1\%} & \multicolumn{1}{c}{9.9\%}  \\
\cline{1-7}
\multirow{3}{*}{CMD}&\multicolumn{1}{l}{\multirow{2}{2em}{Video-only}} & Correct & \multicolumn{1}{c}{50.5\%}          & \multicolumn{1}{c}{12.6\%}  & 63.1\% & \multirow{3}{*}{69.9\%}\\
&\multicolumn{1}{c}{}                           & Wrong & \multicolumn{1}{c}{24.0\%}         & \multicolumn{1}{c}{12.9\%}  & 36.9\% \\ 
\cline{2-6}
&\multicolumn{2}{c}{Total} & \multicolumn{1}{c}{74.5\%} & \multicolumn{1}{c}{25.5\%}\\
\cline{1-7}
\multirow{3}{*}{LSMDC}&\multicolumn{1}{l}{\multirow{2}{2em}{Video-only}} & Correct & \multicolumn{1}{c}{90.3\%}          & \multicolumn{1}{c}{2.7\%}  & 93.0\% & \multirow{3}{*}{22.9\%}\\
&\multicolumn{1}{c}{}                           & Wrong & \multicolumn{1}{c}{6.2\%}         & \multicolumn{1}{c}{0.8\%}  & 7\% \\ 
\cline{2-6}
&\multicolumn{2}{c}{Total} & \multicolumn{1}{c}{96.5\%} & \multicolumn{1}{c}{3.5\%} & \\
\bottomrule
\end{tabular}}
\label{tab:human-video-seq}
}
\end{table}

\vspace{0.1in}
\noindent \textbf{Human Annotation and Analysis.} 
We use human annotation to estimate the semantic gap on \shortname{}, CMD, and LSMDC. For each dataset, we recruited 4 human annotators to label 1,000 pairs of video segments both with and without text. We assigned two annotators to the text + video condition and two to the video-only condition. In each condition, both annotators labeled the same 200 data points to compute inter-rater agreement, see Table. \ref{tab:cohen_kappa}. Note that the Cohen's Kappa for CMD is lower than LSMDC and \shortname{}. This is because human annotators found it difficult to identify the order of a significant number of video pairs in CMD and resorted to random guessing. 
\begin{table}
\caption{Cohen's Kappa for \shortname{}, CMD, and LSMDC.}
\centering
\begin{tabular}{llll}
\toprule
&\shortname{} &  CMD & LSMDC \\
\midrule
Text + Video & 87.9\%    & 61.7\%  & 90.5\%    \\
Video Only   & 82.7\%    & 66.3\%  & 86.9\%    \\
\bottomrule
\end{tabular}
\label{tab:cohen_kappa}
\end{table}

We show the results in Table \ref{tab:human-video-seq}. In the video-only condition, the human annotators are correct 82.8\% / 63.1\% / 93\% of the time; when additional text is available, they achieve an accuracy of 90.1\% / 74.5\% / 96.5\%. On 2.7\% / 12.9\% / 0.8\% of the data, annotators in both conditions made the wrong decision.

We conduct a probabilistic analysis to quantify the cross-modality semantic gap. First, we assume that, when humans do not know the answer, they can make random guesses that are 50\% correct; knowing the answer, they always answer correctly. Thus, we need to distinguish knowing the correct answer from making the correct decision. We let $K_{text}$ denote the humans in the text-video condition know the right answer and $K_{video}$ denote the humans in the video-only condition know the right answer. Further, we assume that random guesses are independent. We want to determine if a text provides grounding to the corresponding video segments. Note that the effect of grounding is observable only when humans cannot determine the ordering from the video segments alone but can do so with the help of additional text. This is captured by the probability $P(K_{text} \mid \neg K_{video})$. When the video segments already contain order information, the text may still provide grounding but we cannot discern its effect from the experiment. With the above assumptions, we can show that
\begin{equation}
{P_{\shortname{}}(K_{text} \mid \neg K_{video})} = 68.6\%
\end{equation}
Thus, we estimate 68.6\% of the texts provide sufficiently accurate descriptions of both video segments, from which sequencing decisions can be made. Similarly, we calculate the percentage of video grounding in text to be 30.1\% and 77.1\% for CMD and LSMDC respectively.
We refer interested readers to details in  Appendix \ref{appendix:semantic-gap}. 

The statistical analysis largely agrees with our observations. CMD provides one-sentence summary for each 1-3 minute video, leaving much video content not described and a large gap (1 - 30.1\% = 69.9\%) between the two modalities. In comparison, textual descriptions in LSMDC are extremely detailed and closely match the video. The estimated gap of LSMDC (1 - 77.1\% = 22.9\%) is higher than our expectation, but the estimate may have very high variance and the actual semantic gap is likely much smaller. This is because that the gap is computed from the percentage of video segments wrongly ordered in both experimental conditions, which is a meager 0.8\% for LSMDC and may easily be dominated by inevitable measurement noise. Had this value been 0, the estimated gap would be 0. 

Finally, the semantic gap in \shortname{} is neither too large nor too small. We believe the moderate semantic gap reflects the use of storytelling techniques in the data and provides an interesting yet attainable challenge for computational understanding. 



\begin{figure}[t!]
	\centering
   	\includegraphics[width=\linewidth]{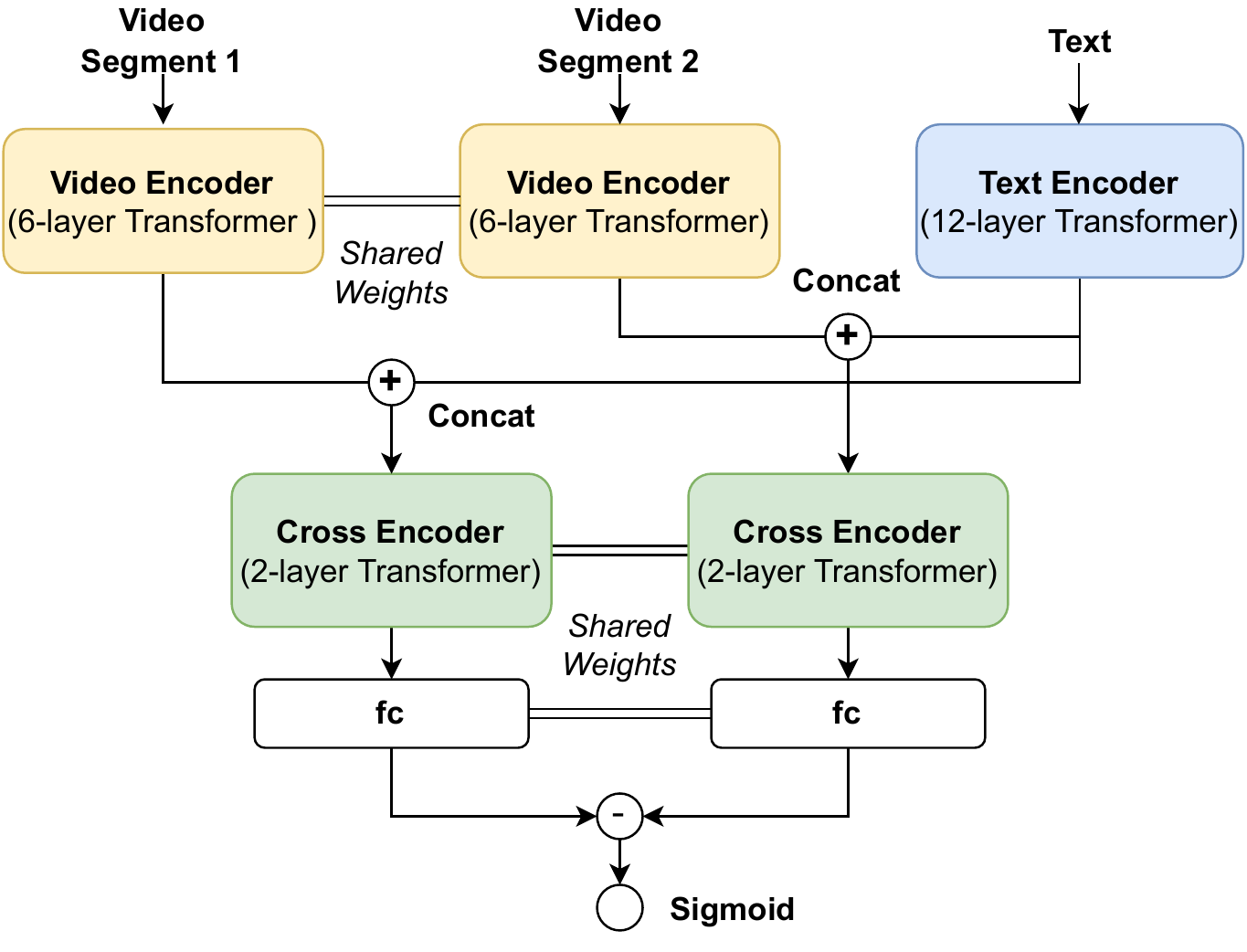}
   \caption{The network architecture for the temporal ordering task. The double vertical lines indicate weight sharing between modules. 
   }
   \label{fig:tempora-ordering-network}
\end{figure}

\vspace{0.1in}
\noindent \textbf{Neural Network Baselines.} 
Similarly to the human annotators, we subject neural networks to two experimental conditions, the video-only and the text + video. We finetune three pretrained modules, the text encoder, the video encoder, and the cross-modality encoder from the UniVL model \citep{luo2020univl}. The two video segments are encoded separately and their features are concatenated with the encoded text feature. After that, the two groups of features go through the cross encoder independently, yielding feature vectors $\bm{f}_1$ and $\bm{f}_2$. With parameter $\bm w$, the prediction is
\begin{equation}
	P(\hat{y}=1)=\sigma(\bm w^\top \bm{f}_1 - \bm w^\top \bm{f}_2),
\end{equation}
where $\sigma(\cdot)$ is the sigmoid function and $\hat{y}$ is the predicted class. Figure~\ref{fig:tempora-ordering-network} shows the overall network architecture. For the network in the video-only condition, we replace the textual feature fed into the cross encoder with an all-zero vector and keep the rest of the network unchanged. 

To cover as much data as possible, we adopt a special dataset split, containing Set A of 2,444 videos, Set B of 2,289 videos, and a validation set of 500 videos. Each network is trained on Set A and tested on Set B, and then trained on Set B and tested on A. We report the average test accuracy. We tuned hyperparameters extensively on the validation set and select the training epoch with the highest validation accuracy. To avoid test data leak, we put all videos of the same movie or movie franchise into the same set. 

We create another baseline from the pretrained Merlot model \citep{zellers2021merlot}, which has been trained to sequence video frames. We extract the center frame from the two video segments and pass them with the corresponding text to the Merlot model, which then predicts which frame comes first. We finetune the Merlot model on \shortname{}. More settings and details can be found in the Appendix.

\begin{figure*}[ht]
	\centering
   	\includegraphics[width=\linewidth]{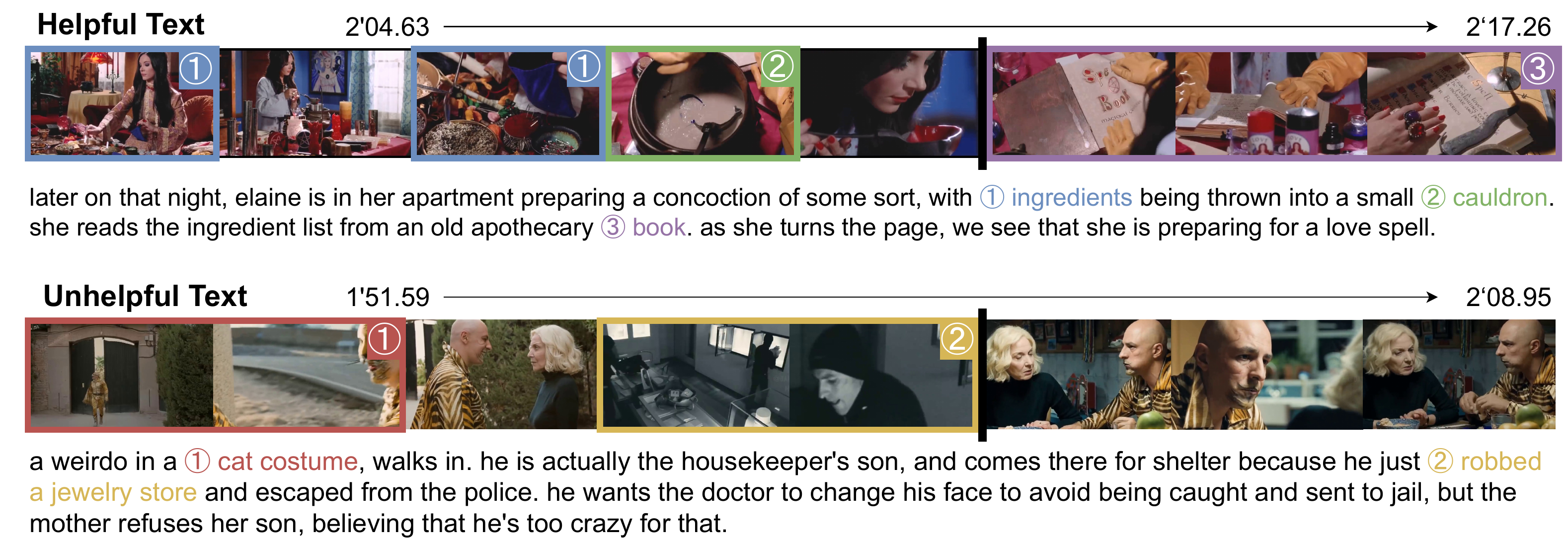}
   \caption{Examples from the most and least helpful text clusters. Bound boxes of the same color in text and video frame denote video-text correspondence. The black line denotes the boundary between the two video segments to be ordered. }
   \label{fig:good_bad}
\end{figure*}

Table \ref{tab:temporal-ordering-accuracy} lists the results. 
The network we trained (UniVL-\shortname) makes correct sequencing decisions $63.4\%$ of the time from video only and $69.1\%$ after incorporating textual information. Merlot achieves $52.6\%$ with both video and text. Both perform substantially below human accuracy, indicating room for improvement.

\begin{table}
\caption{Temporal order prediction accuracy of the text-aware and visual-only models.}
\centering
\begin{tabular}{@{}lll@{}}
\toprule
             & Text + Video & Video Only \\
\midrule
Merlot  & 52.6\%    &    -         \\
UniVL-\shortname              & 69.1\%    & 63.4\%      \\
Human annotator               & 90.1\%    & 82.8\%      \\
\bottomrule
\end{tabular}

\label{tab:temporal-ordering-accuracy}
\end{table}

\vspace{0.1in}
\noindent \textbf{Object and Action Analysis.} 
We examine the match between text and video with object detection and action recognition. First, from the UnivL-\shortname{} sequencing models, we sort texts into helpful and unhelpful based on how much adding text improves the confidence of the correct answer. We take 5\% most helpful texts, which improve the confidence the most, and and 5\% least helpful texts. Meanwhile, we run Faster-RCNN \citep{girshick2015fast} trained on Open Images V4 \citep{kuznetsova2020open} to detect 600 object classes on video frames, and 3D-ResNet \citep{hara2018can} trained on Kinetics-700 \citep{kay2017kinetics} to detect 700 action classes. After that, we match the identified objects and actions to the texts. The Appendix contains more details.

Table \ref{table:object_action} shows that the most helpful texts contain relatively 18.8\% more recognizable objects and 25.0\% more actions than the most unhelpful texts. This suggests that textual references to objects and actions in the video may have contributed to the temporal ordering task. In addition, detected objects and actions appear rather scarcely in the text descriptions, which contain 83 words on average. We  attribute this observation to the semantic gaps explain earlier. 

We further investigate the phenomenon by inspecting two examples (Figure~\ref{fig:good_bad}) from the 5\% most helpful texts and the 5\% least helpful texts, respectively. We observe that the helpful text mentions objects, such as the cauldron and the book, from both video segments, thereby providing ordering information. In comparison, the unhelpful text mentions rare object and action such as cat costume and jewelry robbery, which are difficult for the network to learn. Similarly, matching the text ``the mother refuses her son'' and the discussion shown in video is not straightforward and would require identity tracking and event understanding. 

This case study highlights challenges in understanding realistic videos. It also points to the integration of multiple AI capabilities, including visual recognition on the long tail, long-term identity tracking, and event understanding, as potential future directions.  

\begin{table}
\caption{Number of words that match exactly the detected object names or action names per text description.}
\footnotesize
\centering
\begin{tabular}{@{}p{1.7cm}p{2.0cm}p{2.4cm}@{}}
\toprule
     & Objects Detected & Actions Recognized   \\
\midrule
Unhelpful Text  & 0.16   & 0.16     \\
Helpful Text & 0.19 (+18.8\%)   & 0.20 (+25.0\%)    \\
\bottomrule
\end{tabular}

\label{table:object_action}
\end{table}


\section{Evaluating Story Understanding: Retrieval and Alignment}
\label{sec:retrieval-exp}

In order to evaluate how much computational models understand the multimodal stories of \shortname{}, we need a task that satisfies the following criteria. First, it should be strongly correlated with the level of story understanding that the models can perform. Second, it should be easy to automatically evaluate. Third, it should be relatively easy for humans to annotate, or is compatible with unsupervised learning, so that we can scale the training data. 

We propose that one suitable task is bi-directional text-video retrieval. The task requires retrieving the narration text corresponding to a given video segment and the other way around. Retrieval may seem straightforward, but significant story understanding is required to bridge semantic gaps between the two modalities (as discussed in Section \ref{sec:semantic-gap-categories}). In addition, we surmise that annotating text-video correspondence is easier than motivational states or causal relations. 

In this section, we establish baselines on the task of video-text retrieval on \shortname{} and video-text alignment on YMS \citep{nmatch}, which serve as references for future research. 

\subsection{Approach}
\label{sec:retrieval-network}
\par We employ pretrained text and video encoders from UniVL \citep{luo2020univl}. Due to hardware constraints, we choose not to use the cross-encoder to accelerate training and inference. Empirically, the cross-encoder increases the memory footprint and forces us to decrease the batch size, leading to degraded generalization for retrieval. In addition, the $O(n^2)$ complexity of exact inference, where $n$ is the number of samples in each modality, leads to high computational cost, and the dot product incurs a much lower constant in the $O(n^2)$ complexity than the cross-encoder. 

Following UniVL, we use S3D to extract 1 video feature per second. The video segments are trimmed or appended to 4 seconds. A video segment A is represented as 4 video feature vectors $\{v^{A}_1,\ldots,v^{A}_{4}\}$. The texts are trimmed or appended to 64 tokens. The video features and text tokens are then passed through the video / text encoders. The encoded features for the $i^{\text{th}}$ is denoted as $\bm t_{i}$. The encoded features of the $i^{\text{th}}$ video segment is $\bm v_{i}$. The similarity between the text and the video is simply the dot product of the two vectors.
With randomly sampled negative text features $\bm t_{k}, k \neq i$ and video features $\bm v_{k}, k \neq i$, we use the NCELoss \citep{gutmann2010noise}:
\begin{equation}
\begin{split}
{L}_{\text{NCE}} = & \frac{1}{N} \sum\limits_{i=1}^{N} - \bm{v}_{i}^\top \bm t_{i} 
	+ \log \Big( \exp \bm{v}_{i}^\top\bm t_{i} + \\ & \sum\limits_{k\ne i}^{K}{\exp \bm{v}_{i}^\top\bm t_{k}} + \sum\limits_{k\ne i}^{K}{\exp \bm{v}_{k}^\top\bm t_{i}} \Big),
\end{split}
\end{equation}
where $N$ is the total number of training samples and $K$ the number of negative samples.  Each video segment is matched to the text description spanning its duration. The text is extended on both ends to the nearest sentence boundaries.

One cause for the semantic gaps is long-range references to previous events. To tackle this issue, we implement a long-term video memory mechanism. Briefly, we retrieve the most similar video segment from the past 10 video clips (about 44 seconds). Then, we append the video features and text tokens from the memory video clip to the current video features and text tokens respectively before the video / text encoders.

The similarity function used in memory retrieval attempts to match video segments partially and weigh the similarity with the number of people they share. For each text paragraph associated with the video segments, we perform named entity recognition and extract person names. The similarity between segments $A$ and $B$ is 
\begin{equation}
\label{eq:sim_mem}
\begin{split}
    sim(A, B) = (1+E) \frac{1}{16} \sum_{i=1}^{4}\sum_{j=1}^{4} \cos (v^{A}_i, v^{B}_j),
\end{split}
\end{equation}
where $E$ is the number of shared participants in the two texts and $cos(\cdot)$ is the cosine similarity. Intuitively, $\frac{1}{16} \sum_{i=1}^{4}\sum_{j=1}^{4} \cos (v^{A}_i, v^{B}_j)$ measures visual similarity and considers the contribution of each video pair equally. The $(1+E)$ tracks participants in the video segments, two video segments are more like to be related if the participants overlap.



\subsection{Retrieval on \shortname{}}
The training, validation, and test sets contain 4,191 videos, 500 videos, and 502 videos, respectively. No movies or movie franchises appear in two sets simultaneously. 
The videos are divided into non-overlapping clips, each consisting of two scenes and having a mean duration of 4.4 seconds. YouTube videos often contain introduction and channel information at the beginning and the end, so we exclude 5\% at each end of the videos. 

For evaluation, we report recall at 1, 5, and 10 items (R@1, R@5, and R@10), and Median Rank (MR). As the video and the text are not exactly matched by time, given a video clip, we consider the three closest sentences as correct answers and vice versa. During evaluation, videos of the same movie are removed to avoid noise.

We compare the original UniVL model, the VideoCLIP \citep{videoclip} model without finetuning, UniVL finetuned on \shortname{} data (UniVL-\shortname), as while as UniVL modal finetuned on \shortname{} with long-term memory (UniVL-\shortname{}-memory).

\subsection{Transfer to YMS}
Without in-domain finetuning, we directly test the model trained on \shortname{} on the YMS dataset, which contains 94 YouTube movie summary videos with manual annotation of fine-grained video-text alignment.
To prevent test data leak, we remove any summary videos for the 94 YMS movies from the training set used in this experiment. 

\vspace{0.1in}
\noindent \textbf{Evaluation.} 
In YMS, a text segment may correspond to multiple video clips, whereas a video clip may correspond to one or zero text segments. During inference, we align every video clip to the text segment with the highest similarity, as computed by the neural network. This creates the desired many-to-one alignment. If the highest similarity falls below a threshold, tuned on the validation set, the video clip is considered as not matching anything. 

Following \cite{nmatch}, we use two evaluation metrics. Clip accuracy is defined as the temporal proportion of correctly aligned video segments. Sentence IoU is defined as the intersection-over-union between the aligned video durations and the ground-truth durations. 

\vspace{0.1in}
\noindent \textbf{Baselines.} 
Using the network described in \ref{sec:retrieval-network}, we compare the original UniVL model, the VideoCLIP \citep{videoclip} model without finetuning, UniVL finetuned on \shortname{} data with and without long-term memory (UniVL-\shortname / UniVL-\shortname{}-memory), as well as the supervised NeuMATCH network without the sequential context (\emph{i.e.}, the minimum distance (MD) baseline from \citeA{nmatch}). Details on baseline implementation can be found in the Appendix. Note that UniVL-\shortname{} and UniVL-\shortname{}-memory are trained with two video scenes as the basic unit for retrieval and NeuMATCH-MD uses more finely segmented units. As YMS contains fine-grained annotations, NeuMATCH-MD is trained on in-domain data, and UniVL-\shortname{} and UniVL-\shortname{}-memory are under the transfer learning regime. Thus, this comparison likely puts our network at a disadvantage.  

\vspace{0.1in}
\noindent \textbf{Test Data.} 
For fair comparison with NeuMATCH-MD, we use the original test set of 15 videos and the original video segmentation by \cite{nmatch}, which was performed by a heuristic. In addition, we also create a new split using $70\%$ of the entire YMS as the test set and $30\%$ as the validation set. In this new setting, we also use the new video segmentation strategy in our preprocessing (Section~\ref{sec:segmentation}). The YMS dataset contains sentence-level annotation and sub-sentence annotations. We report performance on both.

\subsection{Results}

\vspace{0.1in}
\noindent \textbf{Retrieval on \shortname{}}
The bi-directional retrieval results are shown in Table \ref{table:retrieval_eval}.  As we expect, the UniVL network finetuned on \shortname{} (UniVL-\shortname) outperforms both the original UniVL and the zero-shot VideoCLIP baseline. Furthermore, the additional of long-term context yielded significant performance gain, with average R@10 improving by 55\%.

\begin{table}
\caption{Retrieval performance on \shortname{} measured using  recall@1, 5, and 10 and median rank. }
\resizebox{\columnwidth}{!}{
\begin{tabular}{@{}p{3.0cm}p{1.0cm}p{1.0cm}p{1.2cm}p{0.9cm}@{}}
\toprule
Model              & R@1 ($\uparrow$)  & R@5 ($\uparrow$)  & R@10 ($\uparrow$) & MR ($\downarrow$) \\
\midrule
\multicolumn{5}{c}{\emph{Text-to-video Retrieval}} \\
 UniVL  & 0.11 & 0.39 & 0.63 & 4818  \\
 VideoCLIP & 0.02 & 0.12 & 0.29 & 3740.5\\
 UniVL-\shortname & 0.76 & 2.30 & 3.58 & 1268  \\
 UniVL-\shortname-memory & \textbf{0.96} & \textbf{3.27} & \textbf{5.41} & \textbf{793}  \\
 \midrule
 \multicolumn{5}{c}{\emph{Video-to-text Retrieval}} \\
  UniVL  & 0.03 & 0.11 & 0.19 & 5687  \\
  VideoCLIP & 0.03 & 0.16 & 0.35 & 4271.5 \\
  UniVL-\shortname & 0.68 & 2.36 & 3.83 & 1304  \\
  UniVL-\shortname-memory & \textbf{1.21} & \textbf{3.92} & \textbf{6.12} & \textbf{755}  \\
\bottomrule
\end{tabular}}

\label{table:retrieval_eval}
\end{table}

\vspace{0.1in}
\noindent \textbf{Transfer to YMS} 
Table \ref{tab:yms1} shows the results of video-text alignment on YMS. Despite the difference in segmentation and the weak supervision from \shortname{}, UniVL-\shortname{} outperforms the supervised NeuMATCH-MD baseline. This shows that UniVL-\shortname{} learns a superior cross-modality distance metric, demonstrating the utility of the large-scale \shortname{} dataset. With sub-sentence level annotations, UniVL-\shortname{} outperforms the original UniVL by $2.6\%$ / $1.7\%$ in the original setting and $2.7\%$ / $0.9\%$ in the new setting. With sentence level annotations, UniVL-\shortname{} outperforms the original UniVL by $2.8\%$ / $2.6\%$ in the original setting and $2.0\%$ / $2.0\%$ in the new setting. Furthermore, UniVL-\shortname{} outperforms zeroshot VideoCLIP and UniVL across all settings. Considering UniVL and VideoCLIP were trained on the gigantic HowTo100M dataset, we attribute the improvement to the similarity between \shortname{} and YMS, which highlights the effectiveness of \shortname{} in the domain of story video understanding. 

Furthermore, long-term context improved average clip accuracy by 18\%. 
We concede that adding memory causes a performance drop in the original split with sentence-level annotation. However, in the original YMS split, only 15 videos are in the test set. Therefore, we believe the results on our new split to be more trustworthy. 

\vspace{0.1in}
\noindent \textbf{Discussion} 
Overall, utilizing long-term memory yielded improvements on both tasks suggesting that the ability to model long-range dependencies is beneficial for understanding videos in \shortname{}. As shown in Section. \ref{sec:semantic-gap-categories}, videos in \shortname{} frequently reference past events at long range. Intuitively, modeling long-range dependencies enable the model to better understand story events. Furthermore, this experiment demonstrates that better story understanding leads to better performance on the retrieval task.  



\begin{table}
\centering
\caption{Alignment performance on YMS. For UniVL-\shortname{}, we report the average and standard deviation from 5 different random seeds.}

\resizebox{\columnwidth}{!}{
\begin{tabular}{@{}lll@{}}
\toprule
                     & Clip Acc. & Sent. IoU     \\
\midrule
\multicolumn{3}{c}{\emph{Original Split (sub-sentence level)}} \\
UniVL     & 3.3  & 1.0           \\
VideoCLIP & 4.8  & 0.6   \\
NeuMATCH-MD (Supervised) & 4.0  & 2.4           \\
UniVL-\shortname & $5.9\pm 0.3$  & $\bm{2.7\pm 0.2}$          \\
UniVL-\shortname-memory & $\bm{6.5\pm 0.3}$ & $ 2.6\pm 0.2 $ \\ 
\midrule
\multicolumn{3}{c}{\emph{New Split (sub-sentence level)}}\\
UniVL     & 7.4  & 1.0           \\
VideoCLIP & 7.6  & 0.7      \\
UniVL-\shortname & $10.1\pm 0.4$  & $1.9\pm 0.1$           \\
UniVL-\shortname-memory & $\bm{13.5\pm 0.3}$ &$\bm{2.6\pm 0.1}$ \\ 
\midrule
\multicolumn{3}{c}{\emph{Original Split (sentence level)}} \\
UniVL     & 4.6  & 0.8           \\
VideoCLIP & 4.0  & 1.1    \\
UniVL-\shortname & $7.4\pm 0.1$  & $\bm{3.4\pm 0.2}$           \\
UniVL-\shortname-memory & \bm{$7.5\pm 0.4$} & $ 2.1\pm 0.2 $ \\  
\midrule
\multicolumn{3}{c}{\emph{New Split (sentence level)}}\\
UniVL     & 5.7  & 1.3           \\
VideoCLIP & 4.9 & 1.0  \\
UniVL-\shortname & $7.7\pm 0.2$  & $\bm{3.3\pm 0.2}$           \\
UniVL-\shortname-memory & \bm{$8.7\pm 0.3$} & $ 3.2\pm 0.2 $ \\ 
\bottomrule
\end{tabular}}

\label{tab:yms1}
\end{table}

\section{Conclusion}
In this work, we present a multimodal story dataset, \shortname{}. We quantitatively characterize the dataset and establish the status of \shortname{} as the largest and most prototypical multimodal story dataset. 
Furthermore, we establish multimodal retrieval baselines for \shortname{} and a zero-shot alignment baseline on YMS to demonstrate the effectiveness of the in-domain \shortname{} in story understanding. We believe that \shortname{} will serve as a new challenge for the research community and inspire new advances in multimodal learning. The dataset will be released upon acceptance.

\section{Limitations}
We identify several limitations. First, the preprocessing of the \shortname{} dataset relies on several automatic techniques. For example, the text description may be written by the original authors but may also come from Google automatic speech recognition. The punctuation restoration model relies on the text. This pipeline architecture may cause errors to propagate. However, given the amount of data, their complexity, and available research budget, we find it unrealistic to manually annotate everything. 
Fortunately, automatic tools improve over time and the preprocessing can be performed again as better tools become available. We opt to make the dataset available now rather than waiting until perfection. 

Second, the human annotators in the video sequencing task do not see the entire story video when making decisions. We choose this setting because the story videos are quite long and omitting them simplifies the annotation task without sacrificing predictive accuracy too much. However, this may have lowered our estimates of human performance. 

Third, the English-only composition is a limitation of \shortname{}. In the next iteration, we plan to expand the dataset to include other languages. 

Fourth, the movies and TV-shows featured in \shortname{} comes predominately from western producers. Therefore the depiction of cultural and social conventions in \shortname{} may be biased towards a western perspective. 

\bibliography{sn-bibliography}

\appendix

\section{Story Coverage}
\label{appendix:coverage}
\noindent \textbf{Dynamic Time Warping.} 
We present the formal definition of the DTW problem: given the WikiPlots sequence of sentences $A=(a_1, \ldots, a_N)$ and the video narration sentences $B=(b_1, \ldots, b_M)$, we seek the best set of correspondence $\{(a_i, b_{g(i)})\}_{i=1}^N$, where the function $g(i) \in \{\epsilon, 1, \ldots, M\}$ returns the index in sequence $B$ that matches sentence $a_i$ in $A$. Further, we introduce a special symbol $\epsilon$. Setting $g(i) = \epsilon$ signifies that $a_i$ is not matched with any sentence in $B$. 

The DTW algorithm can be understood as finding the shortest path in a graph, where each node $(i,j)$ in the graph represents matching sentence $a_i$ and sentence $b_j$. 
The graph contains dummy nodes $(0, 0)$ and $(N+1, M+1)$. From node $(i, j)$, we can transit to node $(i+1, j+1)$, which would match $a_{i+1}$ with $b_{j+1}$ and incur cost $c(i+1,j+1)$.
\begin{equation}
    c(i+1,j+1) = 1 - P(a_{i+1} \Leftrightarrow b_{j+1}).
\end{equation}
Here $P(a_{i+1} \Leftrightarrow b_{j+1})$ denotes the probability that sentences $a_{i+1}$ and $b_{j+1}$ are equivalent, as determined by the Natural Language Inference classifier. 

Similarly, we can transit from $(i, j)$ to $(i+1, j)$, which would match $a_{i+1}$ with $b_{j}$ and incur cost $c(i+1,j)$. The transition from $(i, j)$ to $(i, j+1)$ is symmetric. Additionally, we can transit from $(i, j)$ to $(i, j+1, \epsilon)$, which prevents $b_{j+1}$ from matching anything. From $(i, j+1, \epsilon)$, we may transit to $(i, j+2, \epsilon)$, $(i, j+2)$, or $(i+1, j+2)$. The costs of ignoring a sentence in $A$ and $B$ are $\delta_A$ and $\delta_B$ respectively. With this setup, the best correspondence can be found as the path from $(0, 0)$ to $(N+1, M+1)$ with minimum cost. 
We use grid search to find optimal $\delta_A$ and $\delta_B$ between $0.1$ and $1.2$ using manually labeled sentence correspondence.

\begin{figure}[t]
	\centering
   	\includegraphics[width=\linewidth]{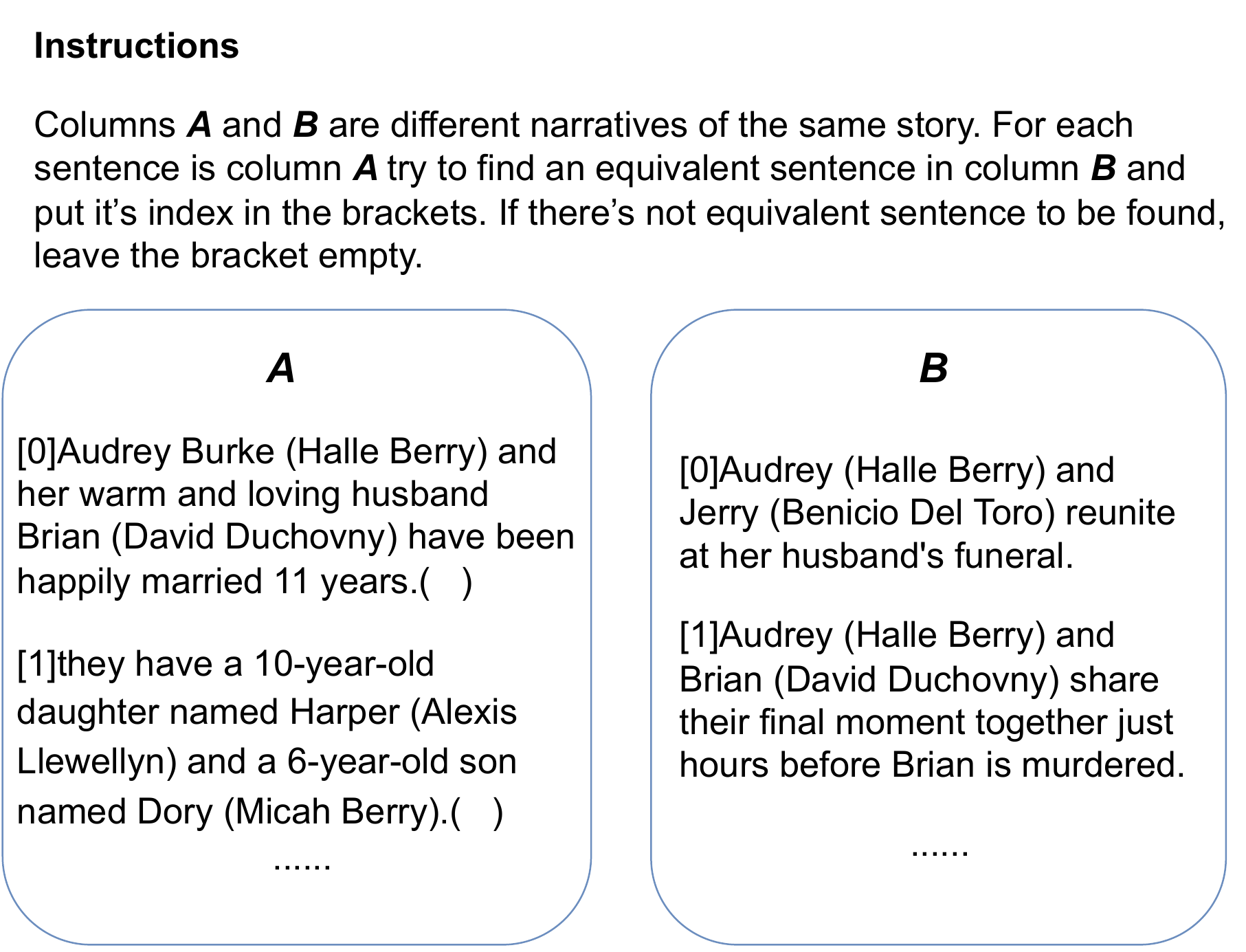}
   \caption{Annotator Instructions for Story Coverage}
   \label{fig:instruction}
\end{figure}

\vspace{0.1in}
\noindent \textbf{Annotation instructions.} Figure~\ref{fig:instruction} shows the instructions we give to our annotator. Here column $A$ is the WikiPlot summary and column $B$ is the summary from \shortname{} or CMD or LSMDC.

\vspace{0.1in}
\noindent \textbf{Justification for the size of labeled validation set.}
To find the optimal hyperparameter for DTW, we manually label 1000 sentences from CMD and \shortname{}, and 2000 sentences from LSMDC. 

Although the labeled set is small compared to the size of the dataset, we believe it is sufficient for the purpose of the hyperparameter search, as adding more annotated examples does not change the results significantly. We conduct another experiment with 500 labeled sentences from CMD and \shortname{}, and 1300 sentences from LSMDC using exactly the same experimental protocol and symmetric entailment for sentence similarity.
\begin{table}
\caption{Estimated story coverage with different validation set sizes. The display is in the form of CMD validation size / LSMDC validation size / \shortname{} validation size}
\centering
\begin{tabular}{@{}llll@{}}
\toprule
Validation Set Size & CMD     &  LSMDC &\shortname{} \\
\midrule
1000 / 2000 / 1000 & 10.6\%  & 18.1\% & 44.3\%\\
500  / 1300 / 500  & 10.8\%  & 18.1\% & 37.9\%\\
\bottomrule
\end{tabular}

\label{tab:story-coverage-size}
\end{table}

Table. \ref{tab:story-coverage-size} shows the results. Increasing the validation set size from 500/1300/500 to 1000/2000/1000 causes little change in the calculated story coverage. Thus, we conclude that the gain from increasing the labeled validation set size has saturated. 

\section{Mental-state Description}
\label{appendix:mental-state}
\noindent \textbf{Commentary Labeling Protocol.} 
To estimate the amount of commentary text in \shortname{} and determine if the high frequency of mental-state description is due to narrator commentary, we recruited workers from AMT to label the sentences as either commentary or non-commentary. We organize the sentences into sets of five. The workers were instructed to read the five sentences and identify sentences that they consider commentary. To ensure quality, we only recruited AMT workers with a ``Mechanical Turk Master'' qualification and assign at least 3 workers to each set of sentences. We consider a sentence as commentary if 2 or more workers agree that it is commentary. As a verification mechanism, we manually labeled a small test set of 100 sentences. The above AMT labeling protocol achieved 100\% accuracy on this test set.

\begin{figure*}[t]
	\centering
   	\includegraphics[width=\linewidth]{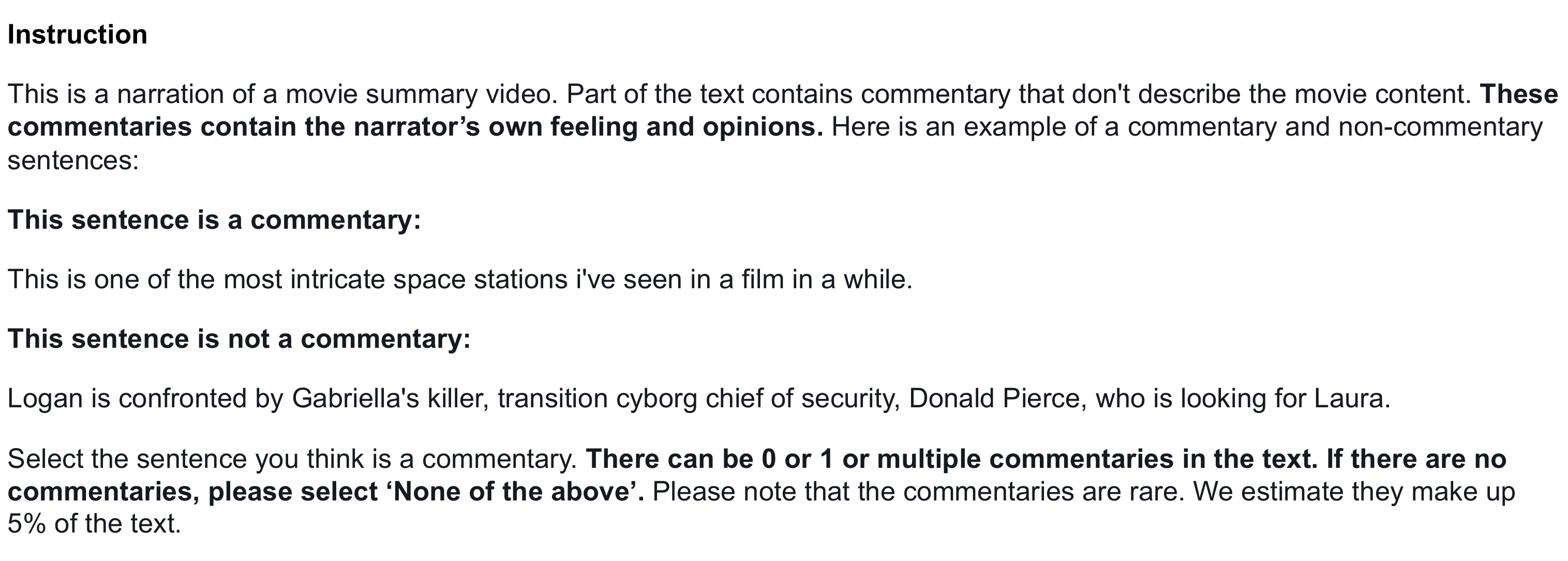}
   \caption{Annotator Instructions for Commentary Labeling}
   \label{fig:instruction_mental_state}
\end{figure*}

Figure~\ref{fig:instruction_mental_state} shows the annotation instruction given to the AMT workers along with a set of five sentences.

\section{Estimating the Semantic Gap}
\label{appendix:semantic-gap}
First, we introduce the following notations. $C_{text}$ denotes that the human under the video + text condition provided the right answer, and $C_{video}$ denotes that the human with video only provided the right answer. Alternatively, $\neg C_{text}$ and $\neg C_{video}$ denote the situations where they did not get the right answer respectively. 

We reckon that the human labels may be the results of random guessing and would like to differentiate random guesses from actually knowing the right answer. For this reason, we introduce notations $K_{text}$ and $K_{video}$, which denote the humans with and without the texts know the right answer respectively. $\neg K_{text}$ and $\neg K_{video}$ denote that the human in those conditions do not know the right answer, respectively. 

\begin{figure}[t]
	\centering
   	\includegraphics[width=0.4\linewidth]{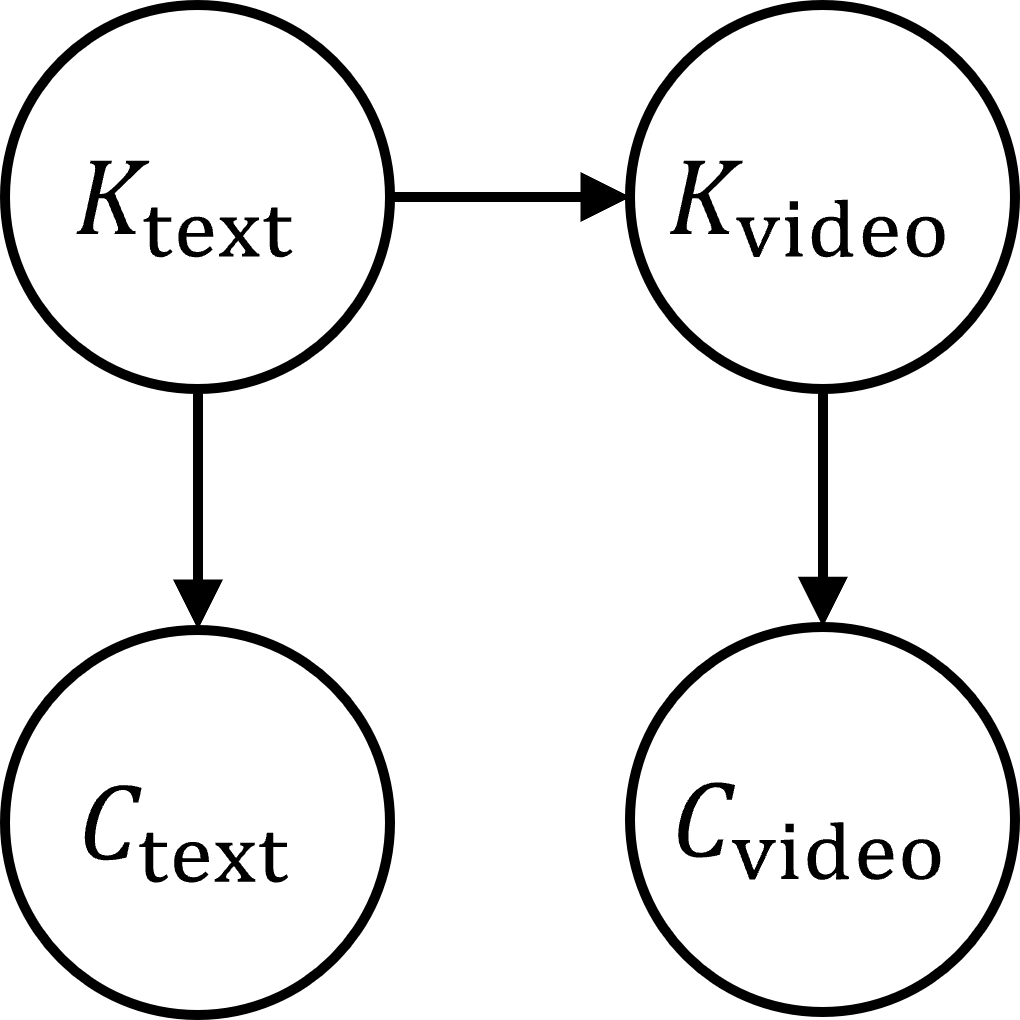}
   \caption{The directed probabilistic graphical model for analyzing user responses. }
   \label{fig:pgm}
\end{figure}

We further assume that, even without knowing the answer, the human can guess 50\% correctly. That is, 
\begin{equation}
\begin{split}
P(C_{text} | \neg K_{text}) & = 0.5, \\  
P(\neg C_{text} | \neg K_{text}) & = 0.5, \\  
P(C_{video} | \neg K_{video}) & = 0.5, \\
\text{and} \, \, P(\neg C_{video} | \neg K_{video}) & = 0.5. \\
\end{split}
\end{equation}
But knowing the answer, the human always makes the correct choice,
\begin{equation}
\label{eq:assumption-2}
\begin{split}
P(C_{text} | K_{text}) & = 1, \\  
P(\neg C_{text} | K_{text}) & = 0, \\  
P(C_{video} | K_{video}) & = 1, \\
\text{and} \, \, P(\neg C_{video} | K_{video}) & = 0. \\
\end{split}
\end{equation}
Finally, we assume the independence relations embedded in the probabilistic graphic model in Figure~\ref{fig:pgm}. In particular, given $K_{text}$ and $K_{video}$, $C_{text}$ and $C_{video}$ are independent. That is, random guesses are independent, but whether humans know the answers in the two conditions is not independent. Given $K_{text}$, $K_{video}$ and $C_{text}$ are independent, and vice versa. 

As an example, we show the calculation of the semantic gap on \shortname{}, the other two datasets can be calculated similarly. With the above independence relations, we have
\begin{equation}
\label{eq:fatoriation}
\begin{split}
& P(\neg C_{text} \wedge \neg C_{video} \wedge \neg K_{video} \wedge \neg K_{video}) \\
= & \, P(\neg C_{text} | \neg K_{text}) P(\neg C_{video} | \neg K_{video}) \\
& \, P(\neg K_{text} \wedge \neg K_{video}). \\
\end{split}
\end{equation}
We know that
\begin{equation}
\label{eq:sum}
\begin{split}
& P(\neg C_{text} \wedge \neg C_{video}) \\ 
= & \, P(\neg C_{text} \wedge \neg C_{video} \wedge  K_{text} \wedge K_{video}) \\
& + P(\neg C_{text} \wedge \neg C_{video} \wedge \neg K_{text} \wedge K_{video}) \\
& + P(\neg C_{text} \wedge \neg C_{video} \wedge  K_{text} \wedge \neg K_{video}) \\
& + P(\neg C_{text} \wedge \neg C_{video} \wedge \neg K_{text} \wedge \neg K_{video}). \\
\end{split}
\end{equation}
We rewrite Eq. \ref{eq:sum} with factorizations similar to Eq. \ref{eq:fatoriation} and eliminate the zero terms due to Eq. \ref{eq:assumption-2},
\begin{equation}
\begin{split}
& P(\neg C_{text} \wedge \neg C_{video}) \\ 
= & \, P(\neg C_{text} | \neg K_{text}) P(\neg C_{video} | \neg K_{video}) \\
& \, P(\neg K_{text} \wedge \neg K_{video}). \\
\end{split}
\end{equation}
Experimental results show
\begin{equation}
P(\neg C_{text} \wedge \neg C_{video}) = 2.7\%.
\end{equation}
Therefore, 
\begin{equation}
\begin{split}
\frac{1}{4} P(\neg K_{text} \wedge \neg K_{video}) = 2.7\%, \\
P(\neg K_{text} \wedge \neg K_{video}) = 2.7\% \times 4 = 10.8\% . \\
\end{split}
\end{equation}
We know from data that
\begin{equation}
P(\neg C_{video}) = 17.2\%.
\end{equation}
From our assumptions, 
\begin{equation}
\begin{split}
P(\neg C_{video}) = & P(\neg C_{video} | K_{video}) P(K_{video}) + \\
& P(\neg C_{video} | \neg K_{video}) P(\neg K_{video}) \\
= & 0 + 0.5 P(\neg K_{video}).
\end{split}
\end{equation}
Therefore, 
\begin{equation}
P(\neg K_{video}) = 2 \times 17.2\% = 34.4\%.
\end{equation}
Putting things together, 
\begin{equation}
\begin{split}
P(\neg K_{text} \mid \neg K_{video}) & = \frac{P(\neg K_{text} \wedge \neg K_{video})}{P(\neg K_{video})} \\
& = \frac{10.8\%}{34.4\%} = 31.4\% \\
\end{split}
\end{equation}
\begin{equation}
\begin{split}
P(K_{text} \mid \neg K_{video}) & = 1 - P(\neg K_{text} \mid \neg K_{video}) \\ 
& = 68.6\% \\
\end{split}
\end{equation}

\section{Computational Video Temporal Ordering}
\vspace{0.1in}
\noindent \textbf{Preprocessing} 
We generate video segments of exactly 8 seconds separated by a hard camera cut at the midpoint. This is done by selecting a camera cut and cutting the video 4 seconds before and after the camera cut. Each video segment is sampled at 16 frames per second and features are extracted with S3D \citep{xie2018rethinking} pretrained on HowTo100M. We extract S3D features every second (i.e. from 16 frames), yielding 8 1024-dimensional video features for each video segment. For video features extraction we use a frame size of $112\times112$. 

We extract the text between the start of the first video segment and the end of the second video segment. To ensure completeness, the text is extended to the nearest sentence boundaries. The maximum number of text tokens is 128. For longer texts, we remove extra tokens from the start and end of the text. For shorter texts, we add zero padding to the end.

\vspace{0.1in}
\noindent \textbf{Training} 
The models are trained for 30 epochs with learning rate warm-up in the first 6 epochs. 
Hyperparameters are tuned manually to achieve the highest accuracy on the validation set.
The text-aware model is trained with a batch size of 128 and learn rate of $5\times 10^{-6}$ and the visual-only model is trained with a batch size of 256 and initial learning rate of $10^{-5}$. We apply cosine learning rate decay and the Adam optimizer to all models. 


\vspace{0.1in}
\noindent \textbf{Merlot baseline.} 
For the Merlot baseline, we initialize with weights trained on the YT-Temporal-180M \citep{zellers2021merlot} dataset. We finetune the Merlot model on \shortname{} for 20 epochs with an initial learning rate of $10^{-4}$ and linear learning rate decay. The finetuning batch size is set to 1024. We use the AdamW optimizer with $\beta_2=0.98$. We apply learning rate warm-up for the first 10\% of training steps.  

\vspace{0.1in}
\noindent \textbf{Calculating overlap between text description and object / action classes.} 
We first tokenize the text description and use part-of-speech tagging to identify nouns and verbs in the text description \citep{bird2009natural}. For matching with object and action detection, we use only the nouns and verbs from the text descriptions and lemmatize them. We also remove common nouns and verbs (``men'', ``women'', ``person'', and ``clothing'' for nouns and ``is'', ``go'', ``to'', ``get'', ``have'', ``look'', ``walk'', ``play'', and ``take'' for verbs) because these are too frequent and provide little information. 
For object detection, we adopt the top 10 class predictions for each clip. For action detection, we divide the segment into scenes and take the top 3 class prediction for each scene.
Finally, we calculate the number of times the nouns or verbs appear in the detected object or action class names.

\section{Retrieval and Alignment}
\vspace{0.1in}
\noindent \textbf{Preprocessing} 
For retrieval, a video segment is extracted every two scenes. For video feature extraction we sample the video at 16 frames per second and use the S3D network \citep{xie2018rethinking} pretrained on HowTo100M to extract one 1024-dimensional feature every 16 frames. The frame size is $112 \times 112$. We extract 4 features from each clip. If the clip is shorter than 4 seconds, zero paddings are added. If the clip is longer than 4 seconds, we only use the first 4 seconds. We take 64 text tokens for each clip with similar zero padding and truncation. 
Text is extracted from between the start and the end of the video clip and extended to the nearest sentence boundaries.

\vspace{0.1in}
\noindent \textbf{Network architecture} 
The video and text encoders consist of 12 and 6 Transformer layers respectively, and are initialized from weights pretrained on HowTo100M. The outputs are then averaged into two 768-dimensional embeddings for video and text. The similarity between a video, text pair is calculated as the dot product of the video and text embeddings.
\begin{figure}[t]
	\centering
   	\includegraphics[width=0.9\linewidth]{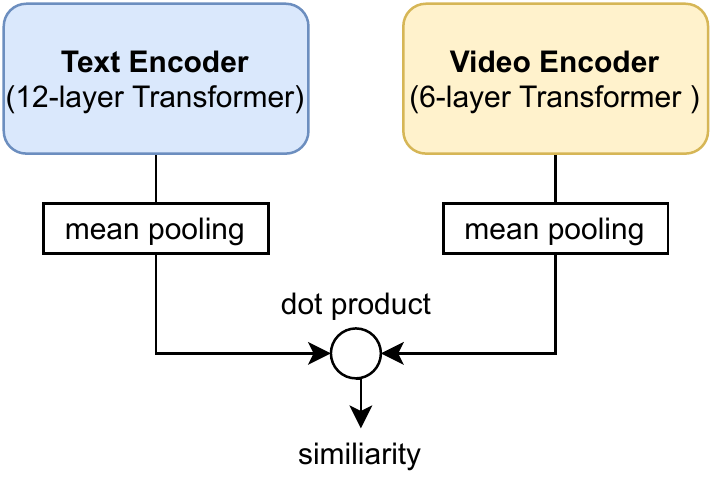}
   \caption{Retrieval Model }
   \label{fig:retrieval-model}
\end{figure}

\vspace{0.1in}
\noindent \textbf{Training} 
The model is finetuned on \shortname{} with an initial learning rate of $5 \times 10^{-5}$ and cosine learning rate decay. We use a batch size of 1024 and train for 20 epochs. The first epoch applies linear warm-up of learning rates. SGD with momentum of 0.9 is used for optimization and a weight decay term of 0.5 is added for 
regularization. The learning rate, learning rate scheduler, weight decay term and batch size are tuned manually to achieve the maximal R@5 on the validation set. Figure~\ref{fig:retrieval-model} shows the model architecture.

\vspace{0.1in}
\noindent \textbf{Baselines} 
For the retrieval experiment, we report results from a single training run finetuned on \shortname{}. For the zero-shot video-text alignment experiment, the UniVL-\shortname{} and UniVL-\shortname{}-memory results reported in the main paper are averaged over 5 training runs. The UniVL and VideoCLIP results are from the UniVL / VideoCLIP models without finetuning. The result of the NeuMatch baseline is directly taken from \cite{nmatch}. 

For the VideoCLIP baseline, we initialize the VideoClip \citep{videoclip} model with weights pretrained on HowTo100M \citep{miech2019howto100m}. VideoCLIP uses S3D network pretrained on 30fps video data as a feature extractor. Therefore during evaluation, the video is sampled at 30 fps. Similar to UniVL, VideoCLIP is used to calculate the similarity between video and text. For retrieval, the similarity is used to retrieve video from text and vice versa. For alignment, a similarity value is computed for each video-text pair in the full-length video, and each video clip is aligned to the text with the highest similarity.

\vspace{0.1in}
\noindent \textbf{Long term memory} 
For any given video segment, we find the most similar video segment from 10 preceding video segments using Eq. \ref{eq:sim_mem} and append its video features and text tokens to the current video features and text tokens respectively. As a result, the input to the video encoder is $8 \times 1024$ vector. We limit the input to the text encoder to 128 tokens. 

The model training is the same as training without memory, except that 60\% of the video and text tokens are masked during training as a form of data augmentation. 

\section{Example Results and Videos}
\begin{figure}[t]
	\centering
   	\includegraphics[width=\linewidth]{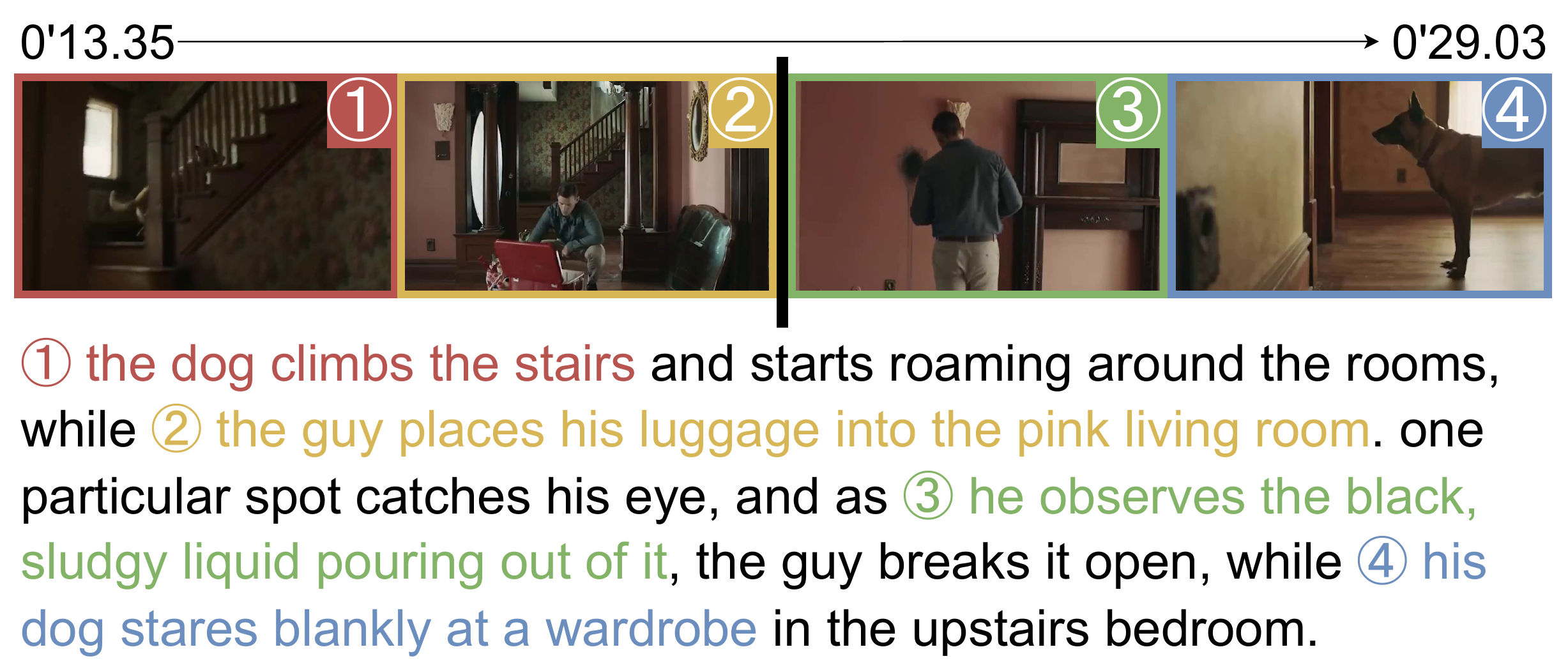}
   	\vspace{0.1in}
   	\includegraphics[width=\linewidth]{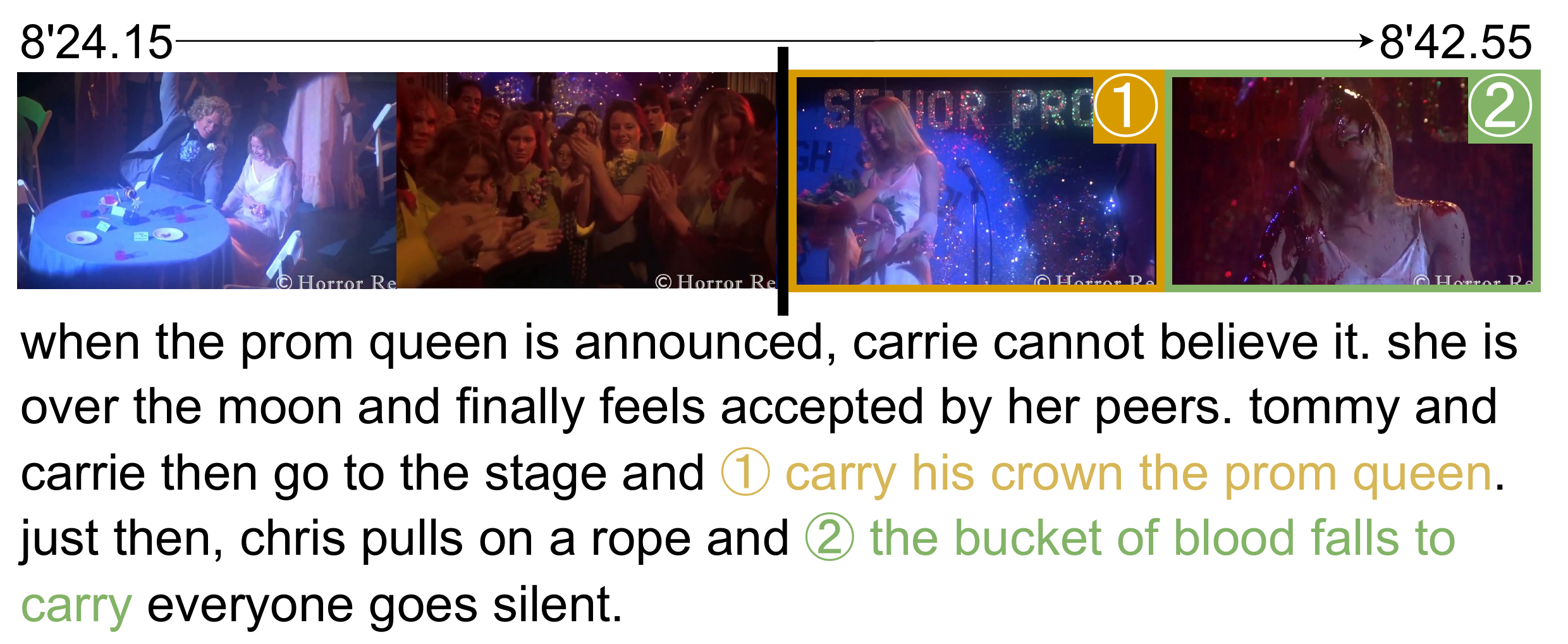}
   	\vspace{0.1in}
   	\includegraphics[width=\linewidth]{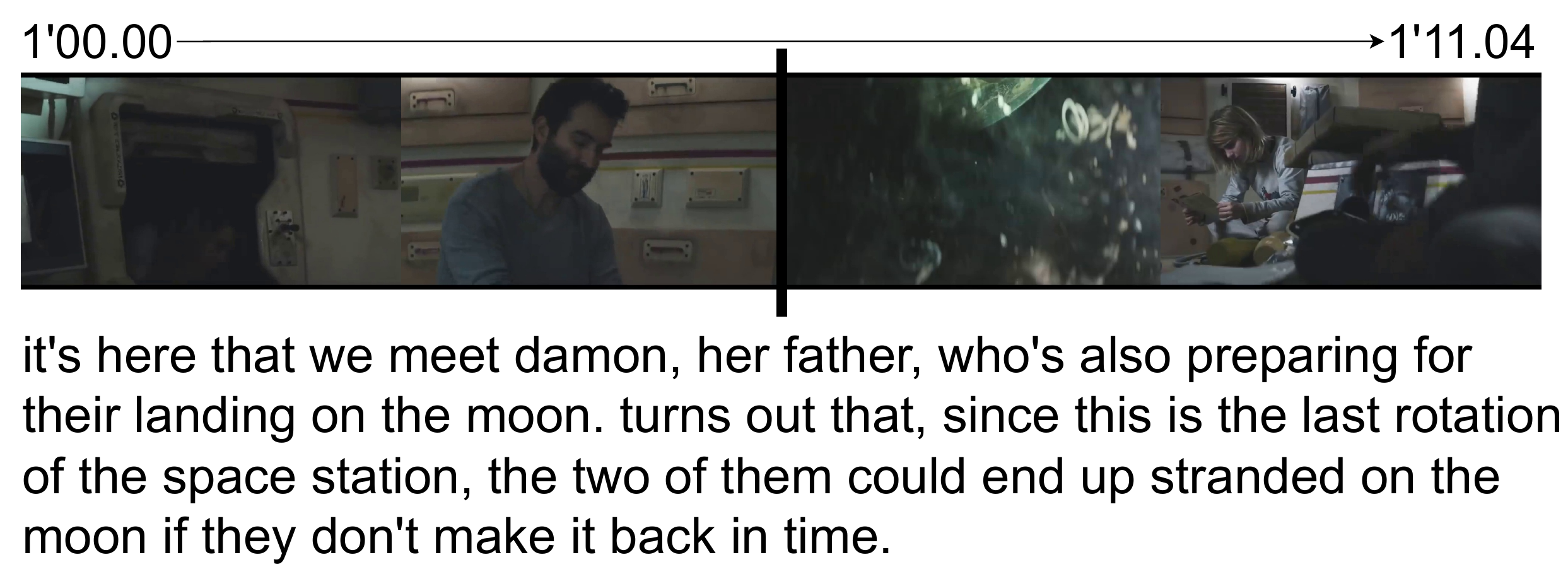}
   	\vspace{0.1in}
   	\includegraphics[width=\linewidth]{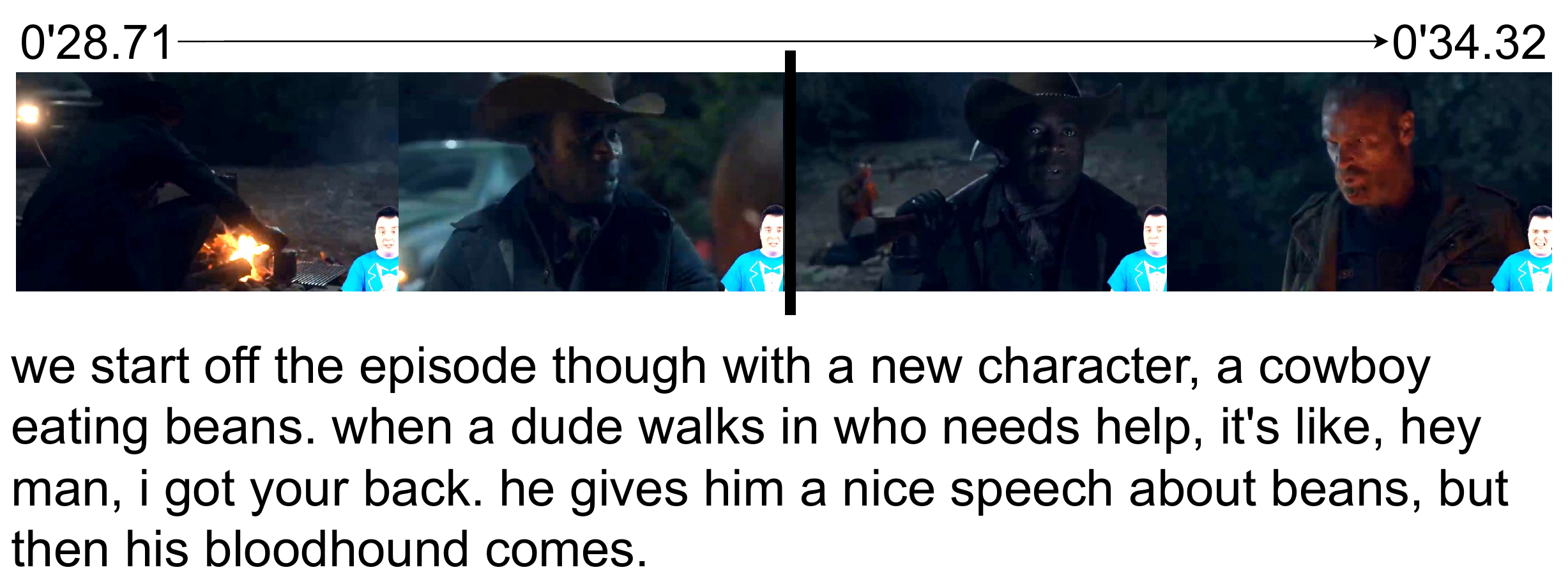}
   \caption{Examples of video segments and text descriptions from \shortname. From the top to the bottom, the videos are summaries for the movies \emph{Girl on the Third Floor}, \emph{Carrie}, \emph{Prospect}, and the TV show \emph{Fear the Walking Dead}. The frames are cropped to the scale of 9:16 for uniform display. We increase brightness for line 4 for clear display. Bound boxes of the same color in text and video frame denote video-text correspondence. The black line denotes the boundary between the two video segments to be ordered. }
   \label{fig:dataset_example_app}
\end{figure}

\begin{figure*}[!ht]
	\centering
   	\includegraphics[width=\linewidth]{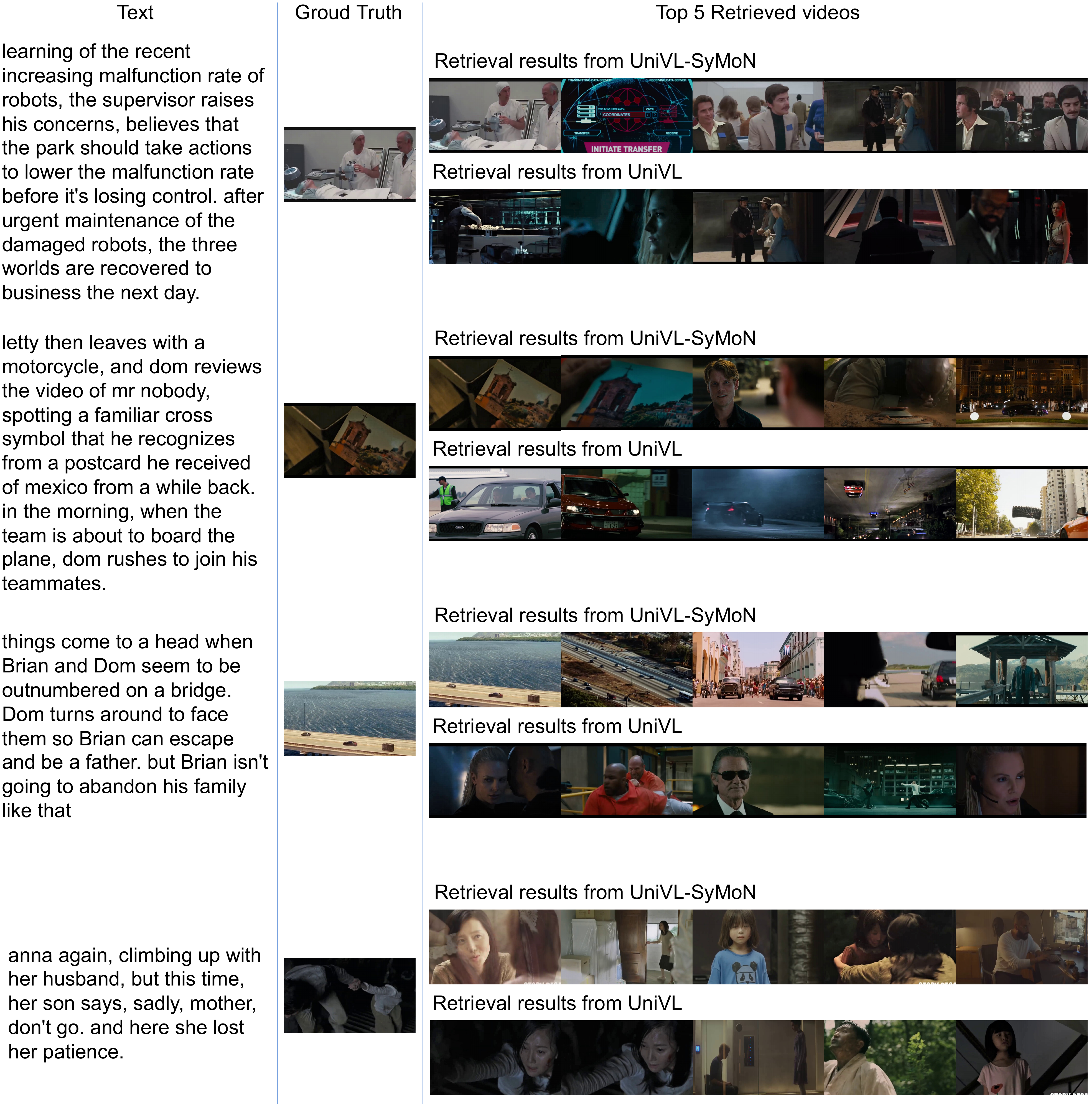}
   \caption{Example of text to video retrieval results with UniVL model and UniVL finetuned on \shortname. The frames are cropped to a constant scale of 9:16, and the brightness of the frames are adjusted for clear display.}
   \label{fig:dataset_example}
\end{figure*}

In Figure~\ref{fig:dataset_example_app} we show some examples from \shortname{}. 
Figure~\ref{fig:dataset_example} shows some example retrieval results from the UniVL and the UniVL+SyMoN models. The first three lines showcases where training on \shortname{} improves retrieval performance. The last line show a case where training on \shortname{} do not improve performance, possibly due to text that are not visually grounded, such as 'her son says, don't go'.

\end{document}